\pdfoutput=1

\documentclass[preprint,12pt,authoryear]{elsarticle}

\usepackage{amssymb}

\usepackage{geometry}
\journal{Journal of \LaTeX\ Templates}
\usepackage{CJK}
\usepackage{indentfirst}
\usepackage{amsmath}
\usepackage{multirow}
\usepackage{colortbl}
\usepackage{subfigure}
\usepackage{diagbox}[2011/11/22]
\usepackage{slashbox}
\definecolor{mygray}{gray}{.92}
\begin{document}

\begin{frontmatter}



\title{Gradient Band-based Adversarial Training \\for Generalized Attack Immunity of A3C Path Finding}


\author[1]{Tong Chen}
\author[1]{Wenjia Niu\corref{cor1}}
\ead{niuwj@bjtu.edu.cn}
\cortext[cor1]{Corresponding author at: Beijing Key Laboratory of Security and Privacy in Intelligent Transportation, Beijing Jiaotong University, 3 Shangyuan Village, Haidian District, Beijing 100044, China.}
\author[1]{Yingxiao Xiang}
\author[1]{Xiaoxuan Bai}
\author[1]{Jiqiang Liu\corref{cor1}}
\ead{jqliu@bjtu.edu.cn}
\author[1]{Zhen Han}
\author[2]{Gang Li}

\address[1]{Beijing Key Laboratory of Security and Privacy in Intelligent Transportation, Beijing Jiaotong University, 3 Shangyuan Village, Haidian District, Beijing 100044, China.}
\address[2]{School of Information Technology, Deakin University.}

\begin{abstract}
  As adversarial attacks pose a serious threat to the security of AI system in practice, such attacks have been extensively studied in the context of computer vision applications. However, few attentions have been paid to the adversarial research on automatic path finding. In this paper, we show dominant adversarial examples are effective when targeting A3C path finding, and design a Common Dominant Adversarial Examples Generation Method (CDG) to generate dominant adversarial examples against any given map. In addition, we propose Gradient Band-based Adversarial Training, which trained with a single randomly choose dominant adversarial example without taking any modification, to realize the ``1:N'' attack immunity for generalized dominant adversarial examples. Extensive experimental results show that, the lowest generation precision for CDG algorithm is $91.91\%$, and the lowest immune precision for Gradient Band-based Adversarial Training is $93.89\%$, which can prove that our method can realize the generalized attack immunity of A3C path finding with a high confidence.
\end{abstract}


\end{frontmatter}


\section{Introduction}
Artificial intelligence (AI) is providing major breakthroughs in solving the problems that have withstood many attempts of natural language understanding, speech recognition, image understanding and so on. The latest studies show that the correct rate of image understanding can reach $95\%$(\cite{he2016deep}) under certain conditions, meanwhile the success rate of speech recognition can achieve $97\%$(\cite{xiong2016achieving}). However, with the development of AI the weaknesses of it are gradually exposed which can be exploited by attackers. Recently, a report(\cite{TRen2017}) issued by Real-time Smart Security Execution Lab of Berkeley University pointed out that there are still many great challenges faces by AI system, and one of the challenges is the adversarial attack.

Adversarial attack was discovered firstly by \cite{Szegedy2013Intriguing} in the context of image classification. Such attacks generally exploit the so-call ``adversarial examples'' to deceive AI systems, which can lead them make mistake. The general form of adversarial examples is the information carrier (such as image, voice or txt) with small perturbations added, which can remain imperceptible to human vision system.

To summarized, the research of adversarial example attack is mainly focused on the following three fields. \emph{In terms of image understanding}, an attacker can construct an adversarial example by adding invisible perturbations on image. Such adversarial example can deceive the automatic driving image recognition system to identify the STOP logo as a speed limit of 60 mark(\cite{kurakin2016adversarial}). Similarity, an attacker can add small perturbations on a panda image to generate an adversarial example, so that the image recognition system can identify it as a gibbon with great confidence(\cite{goodfellow2014explaining}). \emph{In the aspect of speech recognition}, \cite{carlini2016hidden} found that the speaker can emit a noise which human can not recognize as an adversarial example. However, the noise can be correctly identified by certain mobile phones as corresponding voice commands, so as to make the phone switch to flight mode, dial 911 and take other behaviors. Moreover, \emph{in the domain of Atari game}, \cite{mnih2013playing} have also conducted an adversarial example by adding noise to the game background. Which can realize the jamming of the baffle on the ball.

However, few attentions have been paid to adversarial research in the domain of automatic path finding. Due to the universality of robot applications in personal, business or even military, the attack on automatic path finding will generate very serious and unestimated results, which should not be ignored any more. Therefore, in this paper, we would like to take adversarial research on automatic path finding.

In the scenario of path finding, the most important supporting technology is reinforcement learning algorithm. For the pervious research, reinforcement learning algorithm such as Q-Learning(\cite{Konar2013A}), DQN(\cite{Vincent2015Playing}), A3C(\cite{mnih2016asynchronous}), etc. all have been utilized for automatic path finding. Meanwhile, A3C relies on parallel actor-learners and accumulated updates to reduce the training time while improving training stability. Hence, compared with other traditional reinforcement learning algorithms, Asynchronous Advantage Actor-Critic (A3C) is a better and more general RL algorithm for automatic path finding. Therefore, in this paper, we choose A3C path finding as the object for our study.

In the scenario of A3C path finding, adversarial attack is confronted with the ``local map information'', which is different with other existing works in adversarial research. As in the domain of image understanding, speech recognition and Atari game, attackers take adversarial attack to ``global carrier information'' by adding human invisible perturbations to the global image, or generating human indistinguishable speech. We can call adversarial examples that target global carrier information as \textbf{\emph{implicit adversarial examples}}.

However, in A3C path finding, the robot can not obtain the global map image in the process of path finding. In other words, under such scenario, the target of adversarial research is the local map information (such as, environment information for the current position), instead of the global map image. Hence, it is not feasible to conduct an adversarial attack on A3C path finding with implicit adversarial examples.

\begin{figure}
  \centering
  \includegraphics[width=13cm]{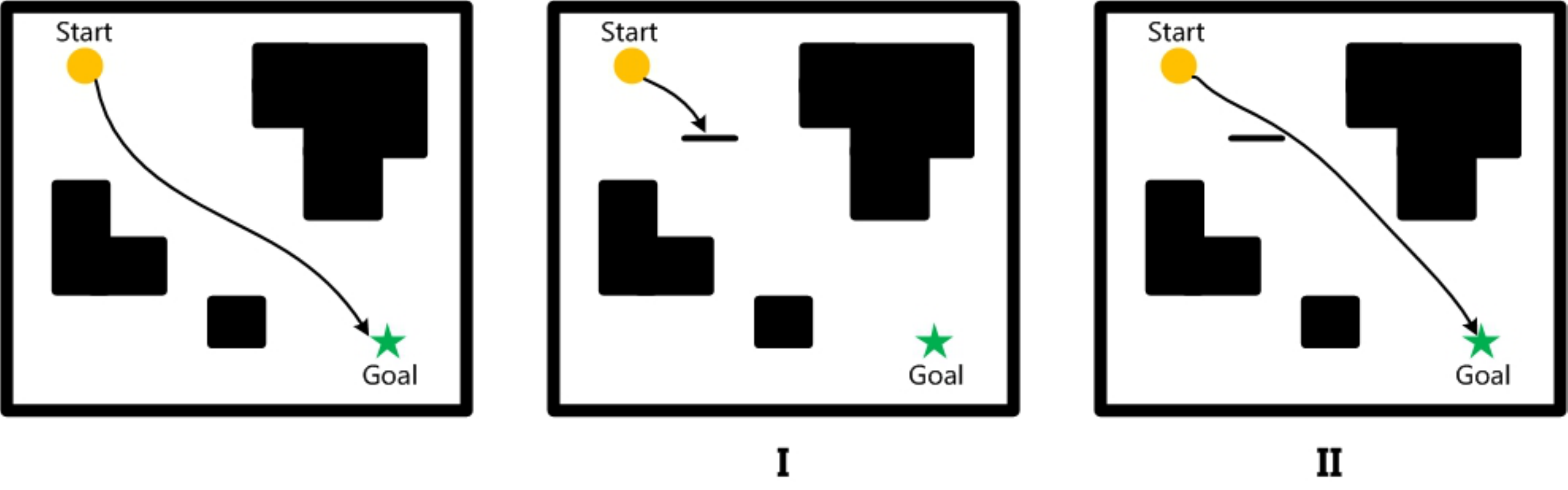}\\
  \caption{The example for results of dominant adversarial attack. The first picture means that for the original map without perturbation, the result of robot automatic path finding based on reinforcement learning algorithm (A3C). We can see that the robot can reach its goal smoothly from the start point. The second picture shows one possible result for successful adversarial attack. For dominant adversarial examples, the robot can not reach the destination successfully. For another possible result, which can be shown in the third picture, robot can reach its goal but the total time is extremely long. It is 10 times or even hundreds times of the normal path finding(which can be shown in the first picture).}
  \label{Fig:resultforattack}
\end{figure}

As there is no well-found work has been proposed to solve this problem to our knowledge. Therefore, in order to fill this gap, we give the very first attempt to design a new type of adversarial example for A3C path finding, which can be called as \textbf{\emph{dominant adversarial example}}. Dominant adversarial example can implement adversarial attack against local map information obtained under A3C path finding, which will make robot can not reach its destination successfully, or let the total time for path finding is extremely long(e,g., Figure~\ref{Fig:resultforattack}). Moreover, experiments have shown that the dominant adversarial example proposed in this paper can realize adversarial attack in A3C path finding successfully with high accuracy.

The main contributions of this paper are as follows:

\begin{enumerate}[(1)]
\item We give the very first attempt to discover dominant adversarial examples in A3C path finding, which can realize adversarial attack successfully. By carrying out the white-box analysis of discovered examples, and summarizing the common rules, we firstly propose a common generation method for dominant adversarial examples, which can generate examples effectively for any given map.
\item We creatively design a defense method, which can be called as ``Gradient Band-based Adversarial Training''.  This method can realize the immunity to attack based on the dominant adversarial example discovered in this paper directly, without modifying the parameters of algorithm or neural network. Moreover, we utilize a single example for Gradient Band-based Adversarial Training, which can realize ``1:N'' attack immunity against generalized dominant adversarial examples.
\end{enumerate}

Therefore, in this paper we realize adversarial research of A3C path finding. We give the very first attempt to discover dominant adversarial examples which can attack successfully, and firstly propose a common generation method for such examples, which can generate examples for any given map. In addition, from the perspective of defense, we design a defense method against generalized dominant adversarial examples, by training with a single discovered example without taking any modification. So as to improve the adversarial robustness of A3C path finding.

The structure of this paper is organized as follow. Section 2 discusses the related works in this field. Section 3 gives a detailed definition of notions and success criteria in this paper. Section 4 describes the overall architecture for Gradient Band-based Generalized Attack Immune Model proposed in this paper. Section 5 gives the experimental evaluation, and the last section gives the conclusion for this paper.
\section{Related Work}
As adversarial attacks pose a serious threat to the security of AI system in practice. This fact has recently lead to a large influx of contributions in this direction. Mainly includes the design of adversarial examples and the proposal of defenses under such attack.

\subsection{Adversarial Examples Design}
In the domain of image recognition system, adversarial attacks have been extensively studied(e.g. \cite{MadryA2017Towards},\cite{MiyatoT2017Virtual}), \cite{kurakin2016adversarial} et al. investigated that the adversarial examples can deceive machine learning in the physical world. Their results showed that 87\% of the adversarial examples can successfully deceive the machine. \cite{LiuY2016Delving} et al. extended the application of adversarial deep learning to image recognition and detection, and they proposed an optimization method named ``ensemble for constructing adversarial pictures''.

In the automatic speech recognition system, \cite{carlini2016hidden} et al. produced adversarial voice commands by adding human indistinguishable noise. The voice commands generated are incomprehensible to the listener but can be interpreted by the device as a correct command.

In Atari game, \cite{mnih2013playing} have also conducted an adversarial examples by adding noise to the game background, which can realize the jamming of the baffle to the small ball.

However, few attention has been paid on adversarial example in automatic path finding. Therefore, under such scenario, we discover dominant adversarial examples which can attack A3C path finding successfully. In addition, we design a common generation method for dominant adversarial examples, based on the white box analysis, which can generate corresponding adversarial examples for any given map.

\subsection{Defense Against adversarial attack}
One defense is to modify the training or input data. \cite{Szegedy2013Intriguing} et al. proposed adversarial training as the first line of defense against adversarial attack. The robustness of network is improved by constantly inputting new types of adversarial examples, and performing adversarial training. Since adversarial training necessitates increased training/data size, is refer to it as `brute-force' strategy. \cite{XieC2017Adversarial} et al. showed that resizing of the adversarial examples can reduces their effectiveness. Moreover, adding random padding to such examples also results in reducing the fooling rates of the networks.

Another way to defense against adversarial attack is to modify the neural networks. \cite{Ross2017Improving} et al. studied input gradient regularization(\cite{DruckerH1992Improving}) as a method for adversarial robustness. Moreover, it has been shown that when combine with `brute-force' adversarial training, can result in a good performance against adversarial attack. \cite{PapernotN2016Distillation} et al. proposed a notion of `distillation' to improve the robustness of deep neural networks against adversarial example, which generated by adding small perturbations to image.

Moreover, some studies proposed utilizing additional networks as a defense against the adversarial attacks. \cite{GaoJ2017DeepCloak} et al. proposed to insert a masking layer immediately before the layer handling the classification. A single training network is added to the original model, which can achieve immunity to adversarial examples without adjusting the coefficients of network. \cite{Xuw2017Feature} et al. proposed to use feature squeezing to detect adversarial perturbation to an image. They utilized two additional models to detect wether the image is adversarial example.

For our work, we creatively design a generalized attack immune model based on a single dominant adversarial example discovered in this paper, instead of modifying the parameters of neural network, or utilizing the additional networks. We can call this model as ``Gradient Band-based Generalized Attack Immune Model''. This model can achieve ``1:N'' generalized immunization against multiple dominant adversarial attacks, by taking adversarial training with a single example. Compared with the traditional adversarial training, we do not need a huge sample expansion for the training set, which can greatly improve the total training speed.

\section{Preliminaries}
In order to solve the problem of adversarial attack under A3C path finding. This paper first propose to utilize dominant adversarial example, which can attack successfully. In this section, we give a detailed definition of implicit/dominant adversarial example respectively. Moreover, we summary the success criteria for dominant adversarial attack in the form of a mathematical formula.
\subsection{Discovery of Dominant Adversarial Example}
To summarize the relevant research works on adversarial attack, we find that the research object of traditional adversarial examples is global carrier information on the pixel level. For example, adding pixel-level invisible noise to the whole image, which can misleads the neural network to give wrong recognition results, or adding pixel-level perturbations to the game background which indistinguishable to human, will make game can not work properly. We can call this kind of example as \emph{implicit adversarial example}, which can be defined as:
\begin{itemize}
\item \emph{\textbf{Definition 1: Implicit adversarial example}} is a modified version of clean information carrier, which generated by adding human invisible perturbations to the global information on pixel level to confuse/fool a machine learning technique.
\end{itemize}

However, under A3C path finding, robot has no idea with the global map image when seeking way on the given map. In the continuous process of unsupervised learning, robot can only perceives the status information of current position, such as, whether obstacles exist in the current position, accessible status of all directions, and whether the robot has reached its destination already. In addition, the robot recognizes the obstacles on physical level instead of pixel-level image information. Therefore, as the processing objects are different (Pixel-level global information/Physical-level local information), utilizing the traditional implicit adversarial examples to attack is ineffective under A3C path finding.

Therefore, we study the adversarial attack in A3C path finding, and discover that \emph{dominant adversarial example} can attack successfully under this scenario, which can be defined as follow:
\begin{itemize}
\item \emph{\textbf{Definition 2: Dominant adversarial example}} is a modified version of clean map, which generated by adding physical-level obstacles to change the local information to confuse/fool A3C path finding.
\end{itemize}

\begin{figure}
  \centering
  \includegraphics[width=15 cm]{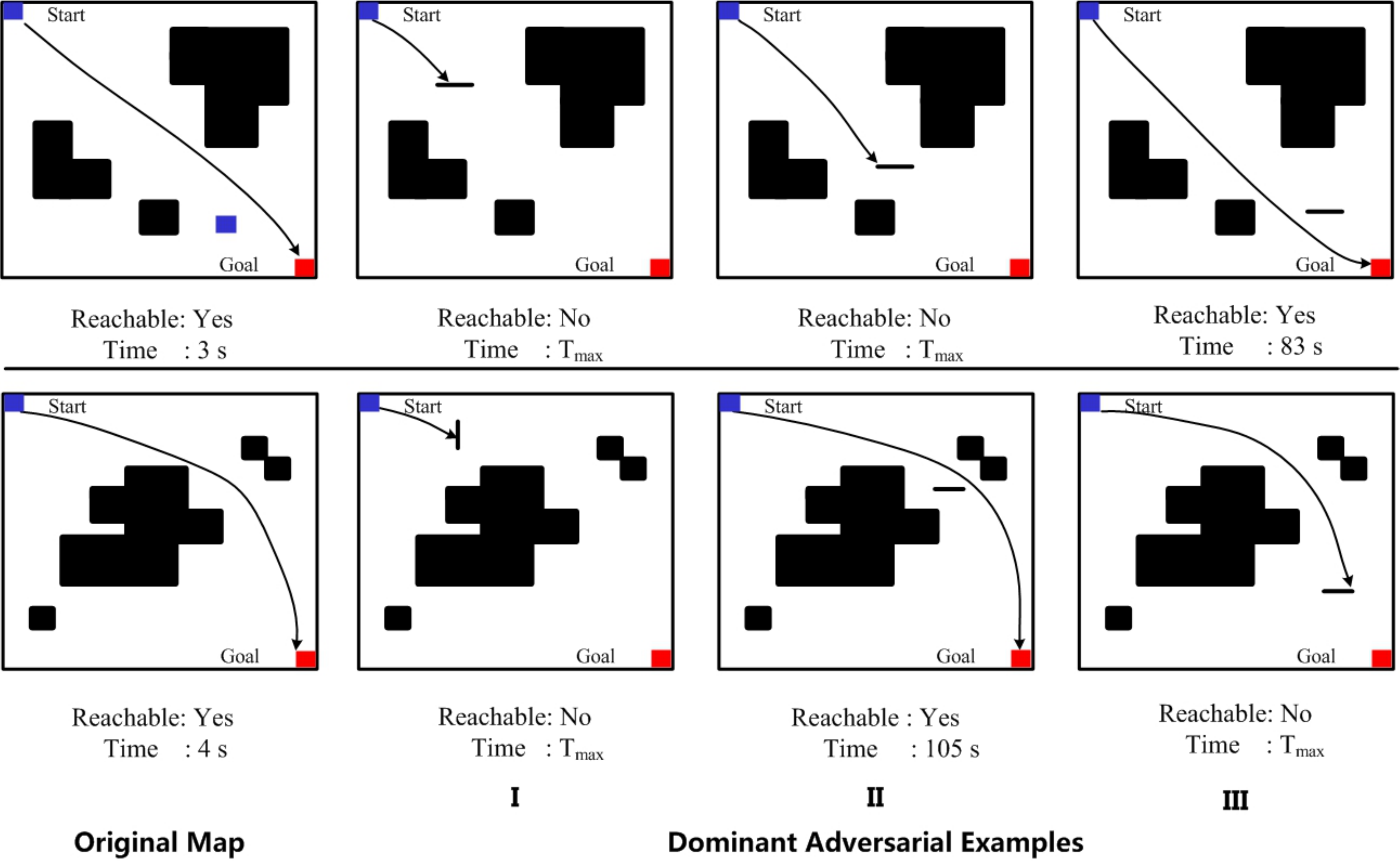}\\
  \caption{The example for dominant adversarial examples in A3C path finding.The first column on the left is the original map without adding any obstacles, and the three columns on the right are examples for dominant adversarial examples after adding perturbations. For each map, we record wether the robot has reached its destination and the total time for path finding. We find that by adding ``baffle-like'' obstacles to the original map can perturb robot path finding effectively.}
  \label{Fig:exampleforDAE}
\end{figure}
Through experiments, we find that adding ``baffle-like'' obstacles to the original map can effectively impact the robot's path finding. As shown in Figure\ref{Fig:resultforattack}, by adding ``baffle-like'' perturbations on the original map, robot can not reach its destination successfully from the start point, or in another case, the robot can achieve the destination but the total time is extremely long compared with the normal situation. It can be seen that the obstacles we added do not block the path of robot at all, in other words, robot can bypass the added obstacles and search map continuously (for example, it can be shown in the experiment that under Q-Learning path finding, robot can bypass the added obstacles and achieve the end point successfully ). However, the robot is still affected by ``baffle-like'' obstacles added and can not find its way properly. As the obstacles we add here is different from implicit adversarial example that invisible to human on pixel level, we call this kind of example as dominant adversarial example(e.g. Figure\ref{Fig:exampleforDAE}), and call attack that utilize such examples as dominant adversarial attack.

\subsection{Success Criteria for Dominant Adversarial Attack}
Meanwhile, in order to determine whether the dominant adversarial example can attack A3C path finding successfully, and to measure the effect of attack, we define the success criteria for dominant adversarial attack in the form of a mathematical formula.

Firstly, we will define two sufficient conditions for the dominant adversarial example success attack:
\begin{itemize}
\item \emph{\textbf{Sufficient condition 1:}} Agent does not reach the target location point eventually.
\end{itemize}

In order to give a mathematical standardization for this sufficient condition, we utilize Euclidean distance(\cite{Danielsson1980Euclidean}) to calculate the distance between agent ending position and the agent target position, which can be defined as follow.
\begin{equation}
  d(A_{actual},D_{destination}) = \sqrt{\sum^n_{i=1}(A_{actual_i}-D_{destination_i})^2}
\end{equation}
where $A_{actual}$ denotes the actual position that the agent eventually arrives, which can be represents as $A_{actual} = \{A_{actual_1},A_{actual_2}\}$, and $D_{destination}$ is the target for robot path finding which can be concluded as $D_{destination} = \{D_{destination_1},D_{destination_2}\}$. If $d(A_{actual},D_{destination}) = 0$ then prove that the agent has achieved the target position, else prove that the attack has succeed, and the greater distance is the better effect of dominant adversarial attack.
\begin{itemize}
\item \emph{\textbf{Sufficient condition 2:}} The total time for agent path finding is too long compared with the normal situation.
\end{itemize}

We utilize the ReLU function to measure the attack effect of dominant adversarial attack. If the total time for path finding is less than the threshold $\varepsilon$ then prove that this attack has failed, and the value for ReLU function will be 0. Moreover, if the total time for path finding is bigger than $\varepsilon$ then prove that this attack has succeed. We can define this condition as follow.
\begin{equation}
  ReLU(T_{total}-\varepsilon)=\left\{
  \begin{aligned}
  ~~~T_{total}-\varepsilon ~~~~~& if~T_{total}-\varepsilon > 0\\
  ~~~0~~~~~~~~ ~~~~~& if~T_{total}-\varepsilon < 0
  \end{aligned}
  \right.
\end{equation}
where $T_{total}$ means the total time recorded for path finding, $\varepsilon$ denotes the threshold for normal time. If $ReLU(T_{total}-\varepsilon) = 0$ then prove that the attack has failed, if $ReLU(T_{total}-\varepsilon) > 0$ denotes that the attack has succeed, and the larger the value is the better effect of dominant adversarial attack.

We stipulate that as long as one or more of the two sufficient conditions is proved to be true, then show that the attack is success. Therefore, the measurement formula for attack effect can be defined as:
\begin{equation}
\begin{aligned}
F_{attack}(A_{actual},D_{destination},T_{total}) &= \omega_1Relu(T_{total}-\varepsilon) + \omega_2d(A_{actual},D_{destination})\\
&= \omega_1Relu(T_{total}-\varepsilon)+\omega_2\sqrt{\sum^n_{i=1}(A_{actual_i}-D_{destination_i})^2}
\end{aligned}
\end{equation}
where $\omega_1$, $\omega_2$ are the weights for two sufficient conditions, and $\omega_2 > \omega_1, \sum\omega_i = 1$. As it can be seen that, the result for robot can not reach its destination is more serious than the increase of pathfinding time. In other words, the first sufficient condition has a greater influence on the attack effect, so its weight in the measurement formula for attack effect should be larger. If $F_{attack} = 0$ then denotes that the attack has failed, if the $F_{attack} > 0$ denotes the attack has been successful, and the greater $F_{attack}$ is, the better effect of dominant adversarial attack.

\section{Gradient Band-based Generalized Attack Immune Model}
\begin{figure}
  \centering
  \includegraphics[width=16cm]{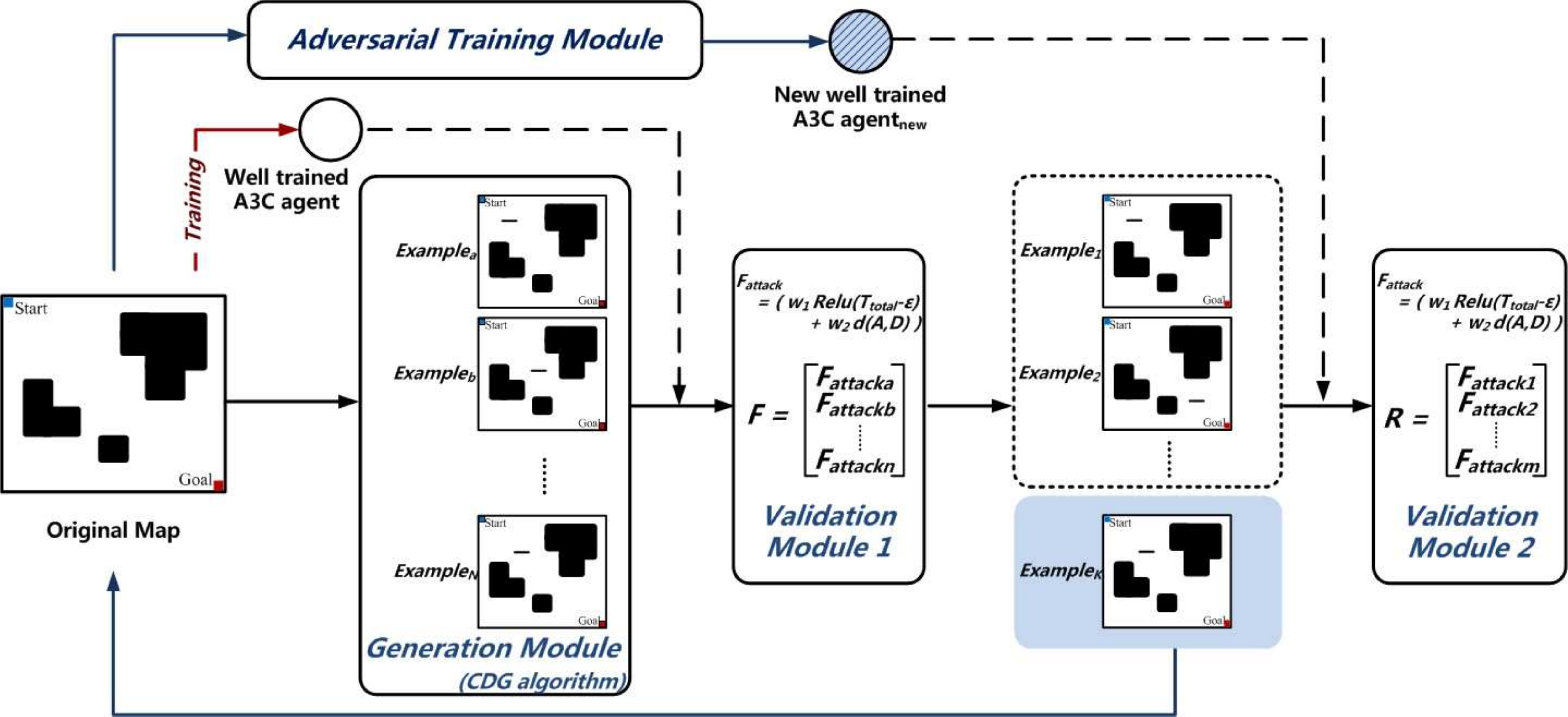}\\
  \caption{Architecture for the gradient band-based generalized attack immune model}
  \label{Fig:architecture}
\end{figure}

In this paper, we propose a generalized attack immune model based on gradient band, which can be shown in Figure ~\ref{Fig:architecture}, mainly consists of \emph{Generation Module}, \emph{Validation Module}, and \emph{Adversarial Training Module}.

For the original clean map,in \emph{Generation Module} we generate dominant adversarial examples $Example=\{Example_1,Example_2,\cdots,Example_N\}$ based on the \emph{Common Dominant Adversarial Examples Generation Method (CDG)} proposed in this paper. Then in the \emph{Validation Module}, we utilize the well trained A3C agent against the original clean map, to calculate the $F_{attack}$ for each example based on the success criteria for attack. Verifying whether the dominant adversarial examples obtained can attack A3C path finding successfully. Selecting the examples whose $F_{attack}>0$, then we can get $Example_{attack}=\{Example_1,Example_2,\cdots,Example_m\}$.

In \emph{Adversarial Training Module}, we utilize a single example $Example_k$ which can attack successfully for adversarial training. Then we can obtain a newly well trained A3C $agent_{new}$. For the last part, we take use of all remaining samples that included in $Example_{attack}$ to attack $agent_{new}$. In the second \emph{Validation Module}, we calculate $F_{attack}$ for each example to verify the immune effectiveness for our model.

\subsection{Common Dominant Adversarial Examples Generation Method(CDG)}
\begin{figure}
  \centering
  \includegraphics[width=15cm]{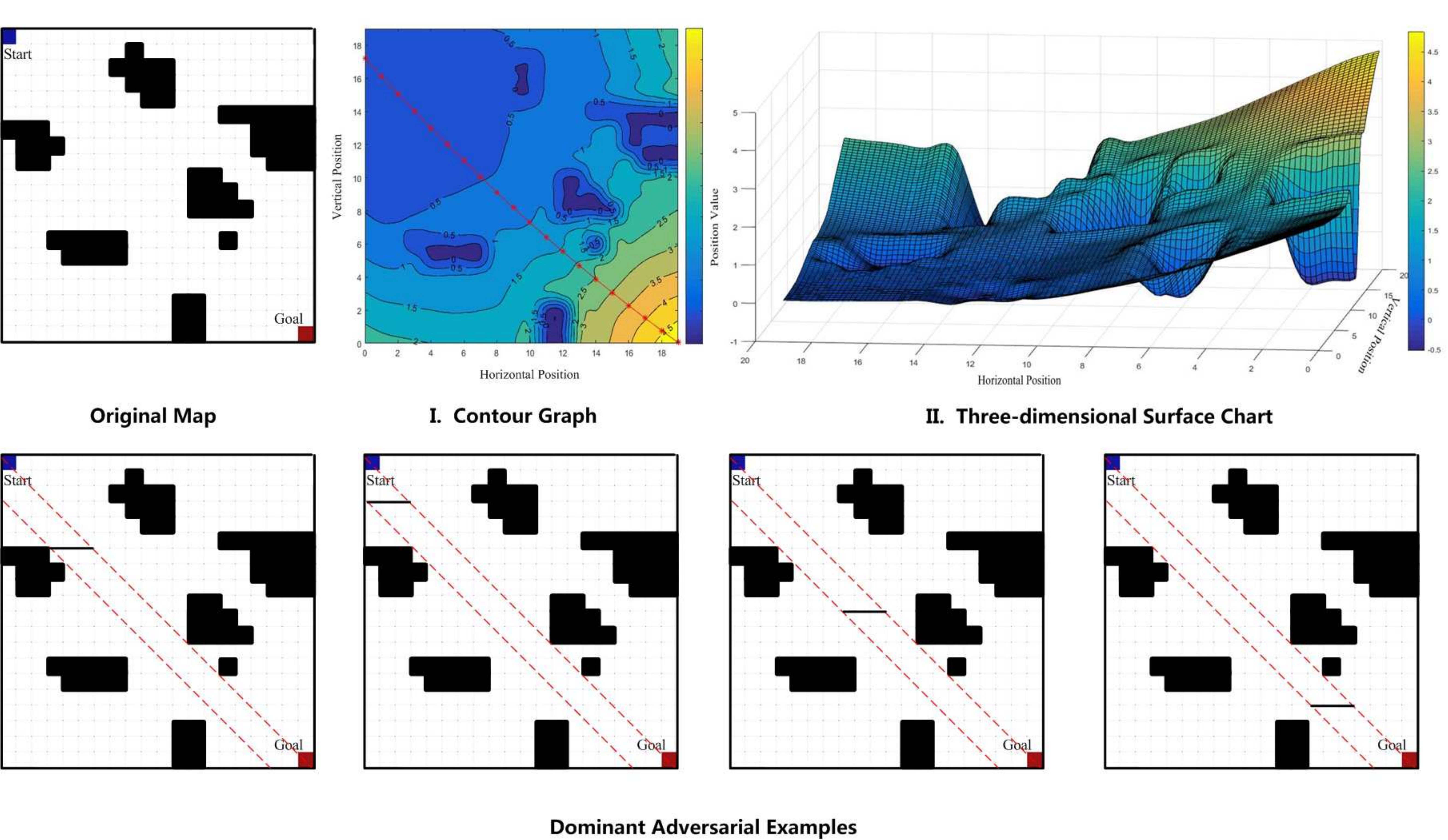}\\
  \caption{The first line shows dominant adversarial examples for the original map. The first picture denotes the original map for attack, and the three columns on the right are the dominant adversarial examples of successful attacks. Meanwhile, the red dotted lines represent the perturbation band. The second line shows the white-box analysis for the value of each position on the original map. The red line denotes the direction in which the value gradient rises the fastest. By comparison between dominant adversarial examples and the contour graph, we find that on the perturbation band, the value gradient rises fastest.}
  \label{Fig:newwhitebox}
\end{figure}

In order to find the generation rule of dominant adversarial examples, we analyze the dominant adversarial examples obtained. As shown in the first line for Figure\ref{Fig:newwhitebox}, are the dominant adversarial examples for the original map which can attack successfully. We find that the "baffle-like" obstacles to these examples all located on a perturbation band, which marked with the red dotted lines. Hence, we give white-box analysis to the well trained agent model for the original map, to get the contour graph, and the three-dimensional surface chart based on the value for each position on the map, which can be shown in the second line for Figure\ref{Fig:newwhitebox}. Moreover, we mark the direction in which the value gradient rises the fastest with red line in the contour graph.

By comparison, we can find that on the dominant adversarial example perturbation band, the value gradient rises the fastest. Therefore, we can also call this perturbation band as ``gradient band''. By adding obstacles on the cross section of gradient band can perturb the agent's path finding successfully. Therefore, we can define the generation rule for the dominant adversarial example as:
\begin{itemize}
\item \emph{\textbf{Generation Rule:}} Adding ``baffle-like'' obstacles to the cross section of gradient band in which the value gradient rises the fastest, can impact A3C path finding.
\end{itemize}

We can utilize the generation rule of dominant adversarial examples to design a common examples generation method (CDG), which can obtain dominant adversarial examples for any given map. As the core for CDG is to add ``baffle-like'' obstacles on the gradient band in which value gradient rises the fastest. Therefore, the first step of this method is to find the most rapid direction of value gradient rises.

\subsubsection{Gradient Function}
In this paper, we take use of Gradient Descent to find the most rapid direction of value gradient rises. The core idea of this algorithm can be concluded as, for each iteration, searching in the negative direction of gradient from the starting point $(X^k, Y^k)$, until we find the optimal value of update step, then we utilize this step to get a new point $(X^{k+1}, Y^{k+1})$ for the next iteration, which can be concluded as:
\begin{equation}
\begin{aligned}
X^{k+1}&=X^k - \alpha_x^k \frac{\partial}{\partial X^k}V(X^k,Y^k)\\
Y^{k+1}&=Y^k - \alpha_y^k \frac{\partial}{\partial Y^k}V(X^k,Y^k)
\end{aligned}
\end{equation}
where V is the value function of X and Y, $\alpha_x^k, \alpha_y^k$ denote the gradient descending step for $X^k, Y^k$ respectively, which need to meet
\begin{equation}
  \phi(\alpha_J^k)=arg minV(J^k-\alpha_J\frac{\partial}{\partial J^k}V(J)), J = X, Y
\end{equation}
in other words, we would like to find $\alpha_J^k$ which can make $X^{k+1}$($Y^{k+1}$) is the point with minimum function value in the gradient descending direction. When the descending height is less than the predefined threshold $\varepsilon$, the drop is stopped and consider that we have obtained the function minimum point.

We can see clearly from the three dimensional surface in Figure\ref{Fig:newwhitebox}, the well trained A3C model has the lowest value near the starting point and the obstacle point, while has the highest value near the ending point. Hence, under such scenario, we need to preprocess the value of each point on the map before using Gradient Descent,
\begin{equation}
  Value(X,Y)_{new} = Value_{max} - Value(X,Y)_{original}
\end{equation}
where $Value(X,Y)_{new}$ and $Value(X,Y)_{original}$ denote the new and original value for point $(X,Y)$ respectively, $Value_{max}$ means the maximum value on the map. Therefore, we transform the problem to find the most rapid direction of value gradient descent for a given map, then we can utilize the Gradient Descent to solve this problem.

\subsubsection{Gradient Band Calculation}
We record the points set $(X,Y)={(x_1,y_1),(x_2,y_2),...,(x_n,y_n)}$ which located on the most rapid direction of value gradient rises, and utilize the Least-Square method to fit a function which can be called as ``gradient function''. We assume that the fitting function is $y = a_0+a_1x+...+a_kx^k$. Then the sum of distances from each point to this curve can be calculated as:
\begin{equation}
  R^2 = \sum^n_{i=1}[y_i-(a_0+a_1x_i+...+a_kx_i^k)]^2
\end{equation}
where $(x_i,y_i)$ denotes the original point, and we can call this step as calculating the sum of deviation squares. In order to find $a_i$ which can satisfies the criteria, we take the partial derivative of $a_i$ to the right side of the equation, then can be represented as a matrix:
\begin{equation}
  \begin{bmatrix}
  n & \sum^n_{i=1}x_i & \cdots & \sum^n_{i=1}x_i^k\\
  \sum^n_{i=1}x_i & \sum^n_{i=1}x_i^2 &\cdots&\sum^n_{i=1}x_i^{k+1}\\
  \vdots&\vdots&\ddots&\vdots\\
  \sum^n_{i=1}x_i^k&\sum^n_{i=1}x_i^{k+1}&\cdots&\sum^n_{i=1}x_i^{2k}\\
  \end{bmatrix}
  \begin{bmatrix}
  a_0\\
  a_1\\
  \vdots\\
  a_k\\
  \end{bmatrix}
  =
  \begin{bmatrix}
  \sum^n_{i=1}y_i\\
  \sum^n_{i=1}x_iy_i\\
  \vdots\\
  \sum^n_{i=1}x_i^ky_i\\
  \end{bmatrix}
\end{equation}
Then to simplify we can get
\begin{equation}
  \begin{bmatrix}
  1&x_1&\cdots&x_1^k\\
  1&x_2&\cdots&x_2^k\\
  \vdots&\vdots&\ddots&\vdots\\
  1&x_n&\cdots&x_n^k\\
  \end{bmatrix}
  \begin{bmatrix}
  a_0\\
  a_1\\
  \vdots\\
  a_k\\
  \end{bmatrix}
  =
  \begin{bmatrix}
  y_1\\
  y_2\\
  \vdots\\
  y_n\\
  \end{bmatrix}
\end{equation}
Therefore, we get the polynomial coefficient matrix A, and finally obtain the gradient function $f(x,y) = y - (a_0+a_1x+...+a_kx^k)$ we need.

In order to calculate the Gradient Band more accurately, we consider two kinds of situations according to the difference for original map and gradient function, one situation is that obstacles exist on both sides of the gradient function, and the other one is that obstacles exist on one side of the gradient function.
\begin{figure}
  \centering
  \includegraphics[width=13cm]{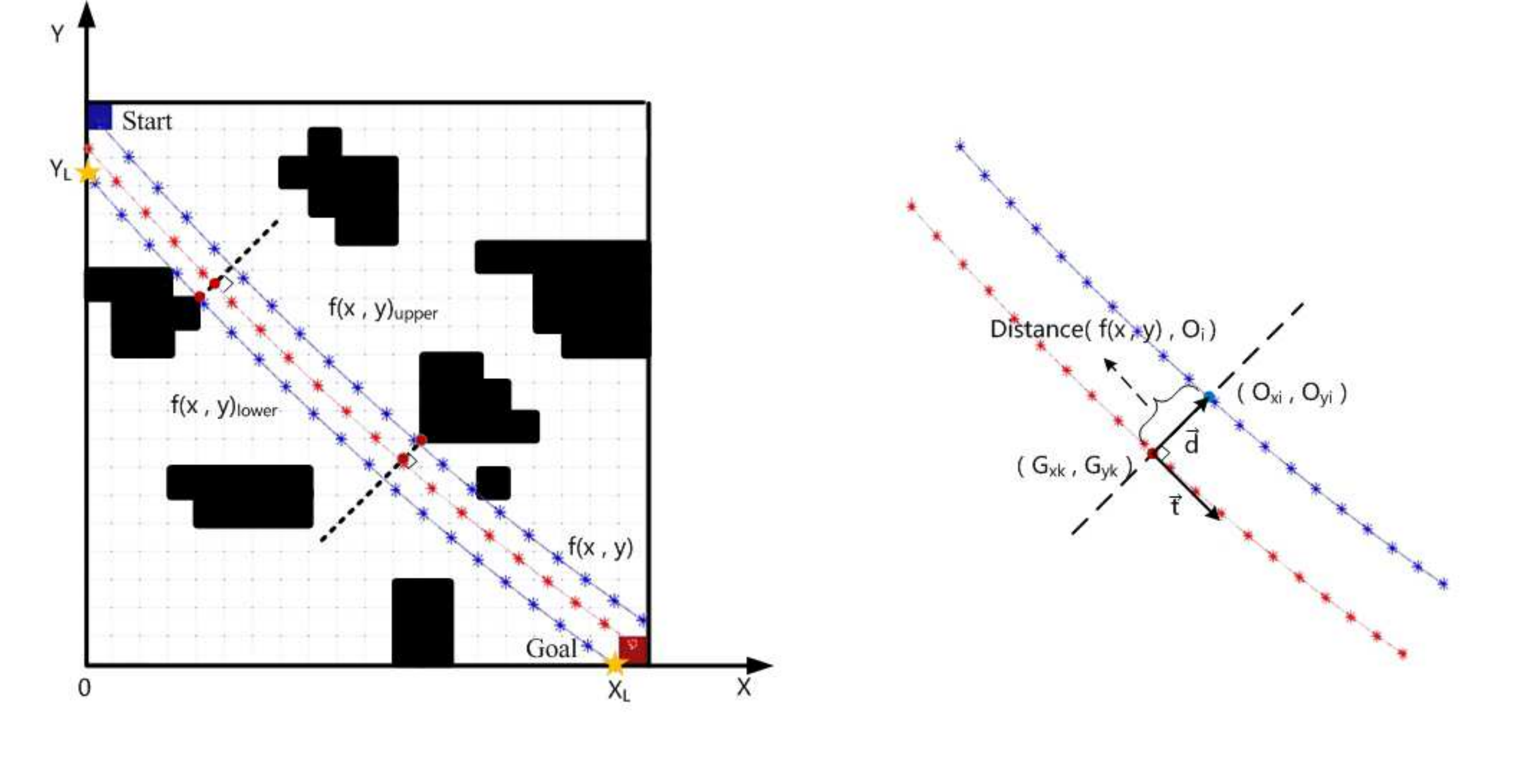}\\
  \caption{Example for Gradient Band calculation under case 1. We establish the coordinate system as shown in the right picture. The red line denotes the gradient function curve $f(x,y)$, the upper blue line $f(x,y)_{upper}$ is the upper bound for the Gradient Band, and the lower blue line is the lower bound for the Gradient Band.}
  \label{gradientbandcase1}
\end{figure}

\textbf{A. Case 1: Obstacles exist on both sides of the gradient function.}

The overall idea for Gradient Band calculation is that, we calculate the distance from all the obstacle edge points in the original map to the gradient function curve, to find the nearest one. Finally, we find the curve which is parallel to the gradient function and passes through this point as the upper/lower bound for the Gradient Band.

Therefore, for the first step, we calculate the distance from all the obstacle edge points $Obstacle = \{(O_{x_1},O_{y_1}),(O_{x_2},O_{y_2}),\cdots,(O_{x_n},O_{y_n})\}$ on original map, to the gradient curve $f(x,y) = y - (a_0+a_1x+...+a_kx^k)$ obtained. Computing the distance from edge point $(O_{x_i},O_{y_i})$ to the gradient function, is to find a point $(G_{x_k},G_{y_k})$ on this curve which can makes the distance vector $\overrightarrow{d} = (G_{x_k} - O_{x_i} , G_{y_k} - O_{y_i})$ perpendicular to the tangent vector $\overrightarrow{t}=(\frac{df}{dG_y},\frac{df}{dG_x})$, which can be shown on the right in Figure\ref{gradientbandcase1}. Then we can get the following equations.
\begin{equation}
  \left\{
  \begin{aligned}
  &f(G_{x_k},G_{y_k}) = 0 \\
  &(O_{x_i}-G_{x_k})\frac{df}{dG_y} - (O_{y_i}-G_{y_k})\frac{df}{dG_x} = 0
  \end{aligned}
  \right.
\end{equation}
By solving this equation group, we can find the nearest point $(G_{x_k},G_{y_k})$ on the gradient curve against the obstacle edge point $(O_{x_i},O_{y_i})$. By calculating the distance between these two points, we can get the distance from $(O_{x_i},O_{y_i})$ to the gradient curve.
\begin{equation}
  Distance(f(x,y),O_i) = \sqrt{(O_{x_i}-G_{x_k})^2+(O_{y_i}-G_{y_k})^2}
\end{equation}

As in this case, obstacles exist on the both sides of the gradient curve, then we need to traverse all the coordinate points in $Obstacle = \{(O_{x_1},O_{y_1}),(O_{x_2},O_{y_2}),\cdots,(O_{x_n},O_{y_n})\}$, and to find the nearest two points from this gradient curve in the upper and lower part respectively. We utilize these two points to calculate the upper and lower bounds of the Gradient Band, which are parallel to the gradient function. Hence, we can conclude the Gradient Band function $F_{GB}(x,y)$ as:
\begin{equation}
  \left\{
  \begin{aligned}
  f(x,y)_{upper}=y-(U + &a_0+a_1x+...+a_kx^k)\\
  f(x,y)_{lower}=y-(L + &a_0+a_1x+...+a_kx^k)\\
  X_L < x < X_{max} &, Y_L < y <Y_{max}
  \end{aligned}
  \right.
\end{equation}
where $f(x,y)_{upper}$ and $f(x,y)_{lower}$ denote the upper/lower bound function respectively, $X_{max}$ and $Y_{max}$ denote the boundary value of the map, $(X_L,0)$ and $(0,Y_L)$ are the intersection points of $f(x,y)_{lower}$ and the coordinate axis, which can be shown in Figure \ref{gradientbandcase1}.
\begin{figure}
  \centering
  \includegraphics[width=13cm]{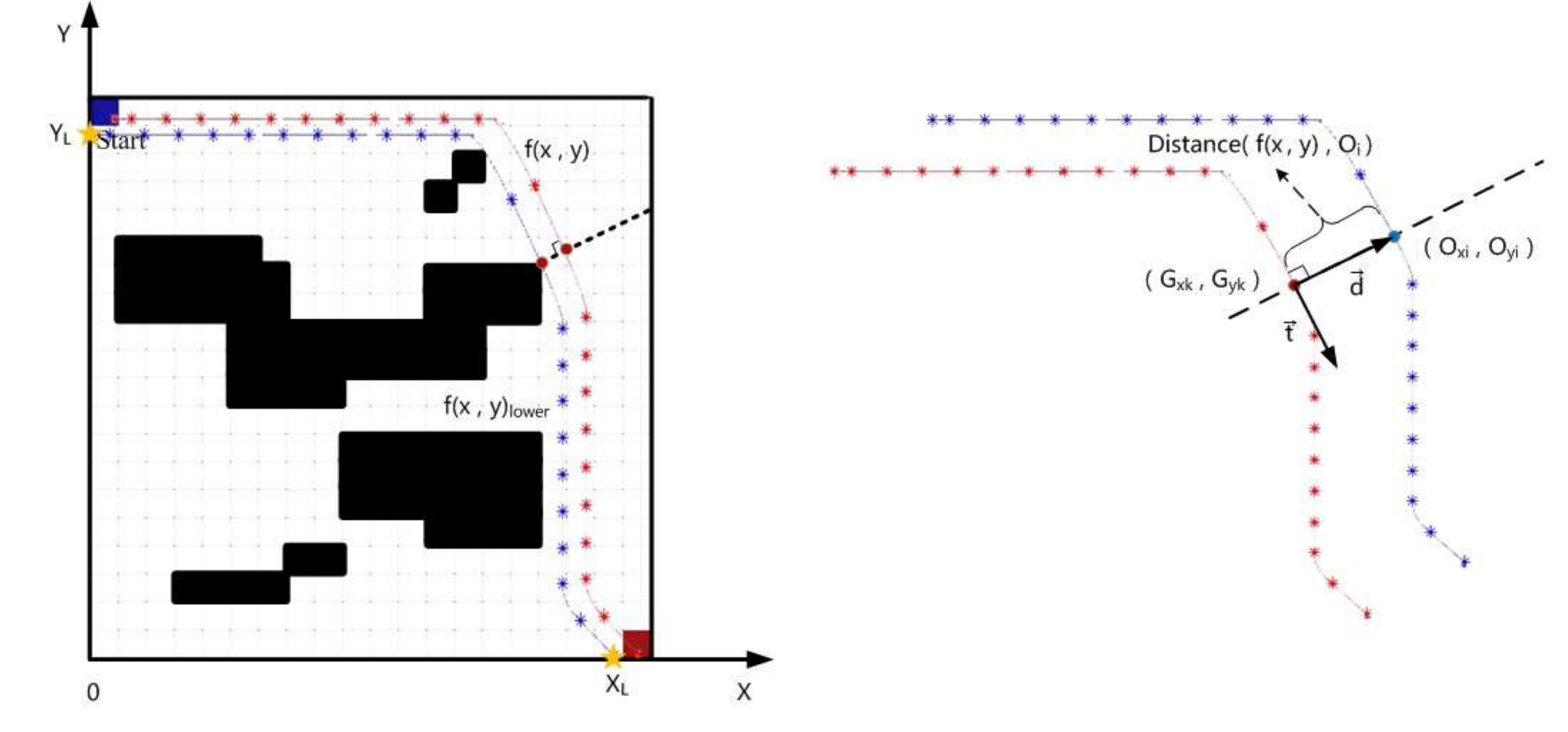}\\
  \caption{Example for Gradient Band calculation under case 2. We establish the coordinate system as shown in the right picture. The red line denotes the gradient function curve $f(x,y)$, the lower blue line is the lower bound for the Gradient Band.}
  \label{Fig:gradientbandcase2}
\end{figure}

\textbf{B. Case 2: Obstacles exist on one side of the gradient function.}

In this case, the calculating for distance between obstacle edge points and gradient function is same with case 1. However, under such scenario, obstacles exist on one side of the gradient function curve, hence, we can only obtain the upper/lower bound function for the Gradient Band. For example, as shown in Figure \ref{Fig:gradientbandcase2}, we can only get the lower bound for the Gradient Band by calculating. Therefore, for this situation, we utilize the boundary point $(X_{max},0)$ and $(0,Y_{max})$ of original map, to get the function equation which is parallel to the gradient function. Then we take the smaller one as the upper bound for Gradient Band. Hence, we can conclude the Gradient Band function $F_{GB}(x,y)$ as:

\begin{equation}
  \left\{
  \begin{aligned}
  f(x,y)_{upper}=\min\{f(&X_{max},0),f(0,Y_{max})\}\\
  f(x,y)_{lower}=y-(L + &a_0+a_1x+...+a_kx^k)\\
  X_L < x < X_{max} &, Y_L < y <Y_{max}
  \end{aligned}
  \right.
\end{equation}

Therefore, for any given map, we can generate dominant adversarial examples by common dominant adversarial examples generation method (CDG), which can be concluded as Algorithm 1.

\subsection{Attack Effectiveness Validation}
In order to verify the attack effectiveness of dominant adversarial examples, we design a validation algorithm which can be concluded as Algorithm 2.

Under different categories of map size, the sample space for dominant adversarial examples is different. For $Map_{N\times N}$, the sample space $S_{Map_{N\times N}}$ can be concluded as:
\begin{equation}
S_{Map_{N\times N}}=
  \begin{bmatrix}
  Example_1\\
  Example_2\\
  \vdots\\
  Example_{2\times (N-1)}\\
  \end{bmatrix}
  ,
  F_{Map_{N\times N}}=
  \begin{bmatrix}
  F_{attack_1}\\
  F_{attack_2}\\
  \vdots\\
  F_{attack_{2\times (N-1)}}\\
  \end{bmatrix}
  ,
V_{Map_{N\times N}}=
  \begin{bmatrix}
  Example_1\\
  Example_2\\
  \vdots\\
  Example_k\\
  \end{bmatrix}
\end{equation}
For each map, we utilize the A3C agent which well trained with the original clean map, to test the attack effectiveness of dominant adversarial examples generated by CDG algorithm in Generation Module. For each example $Map_{N\times N}$ in the sample space, we record the total time for path finding $T_{total_i}$, and the position point that agent actually arrive $A_{actual_i}=\{A_i,Y_i\}$, to calculate the $F_{attack_i}$ value for each example in $S_{Map_{N\times N}}$ to generate $F_{Map_{N\times N}}$. If $F_{attack_i}>0$ then denotes that the $Example_i$ can attack A3C agent successfully, and if the $F_{attack_i}=0$ represents that the $Example_i$ is invalid. Therefore, we can select the dominant adversarial examples which can attack successfully, to generate a valid sample space $V_{Map_{N\times N}}$.
\begin{table}
  \scriptsize
  \centering
  \begin{tabular}{l}
     \hline
     \textbf{Algorithm 1} Common Dominant Adversarial Examples Generation Method (CDG)\\
     \hline
     Input: $Obstacle = \{(O_{x_1},O_{y_1}),(O_{x_2},O_{y_2}),\cdots,(O_{x_n},O_{y_n})\}$,$V(x,y)$ \\
     Output: $O_{baffle}=\{F_{Y_1},\cdots,F_{X_1},\cdots\}$\\
     1. ~~~Gradient function calculation\\
     ~~~~~~~~i. ~~~Processing the value for each point on original map \\
     ~~~~~~~~~~~~~~~~~~~~~~~~~~~~~$Value(X,Y)_{new} = Value_{max} - Value(X,Y)_{original}$\\
     ~~~~~~~ii. ~~~Calculating the fast direction of gradient descent based on Gradient Descent algorithm\\
     ~~~~~~~~~~~~~~~~~~~~~$X^{k+1}=X^k - \alpha_x^k \frac{\partial}{\partial X^k}V(X^k,Y^k)$, $Y^{k+1}=Y^k - \alpha_y^k \frac{\partial}{\partial Y^k}V(X^k,Y^k)$\\
     ~~~~~~iii. ~~~Getting the set of coordinate points for this direction ${(X^1,Y^1),(X^2,Y^2),\cdots,(X^n,Y^n)}$\\
     2. ~~~Gradient Band calculation\\
     ~~~~~~~~i. ~~~Fitting function based on least square method $R^2 = \sum^n_{i=1}[y_i-(a_0+a_1x_i+...+a_kx_i^k)]^2$\\
     ~~~~~~~~~~~~~~~~~~~~~~~~~~~~~~~~~~~~~~$
  \begin{bmatrix}
  1&x_1&\cdots&x_1^k\\
  1&x_2&\cdots&x_2^k\\
  \vdots&\vdots&\ddots&\vdots\\
  1&x_n&\cdots&x_n^k\\
  \end{bmatrix}
  \begin{bmatrix}
  a_0\\
  a_1\\
  \vdots\\
  a_k\\
  \end{bmatrix}
  =
  \begin{bmatrix}
  y_1\\
  y_2\\
  \vdots\\
  y_n\\
  \end{bmatrix}$\\
    ~~~~~~~ii. ~~~Getting the fitting function $f(x,y) = y - (a_0+a_1x+...+a_kx^k)$\\
    ~~~~~~iii. ~~~Gradient Band Calculation\\
    ~~~~~~~~~~~~~~~~A. ~~~Calculating the distance from the obstacle edge points to the gradient function $f(x,y)$\\
    ~~~~~~~~~~~~~~~~~~~~~~~~~~~~~~~~~~~~~~~~~~~~$Distance(f(x,y),O_i) = \sqrt{(O_{x_i}-G_{x_k})^2+(O_{y_i}-G_{y_k})^2}$\\
    ~~~~~~~~~~~~~~~~B. ~~~Finding the obstacle edge points with the smallest distance in the upper/lower parts\\
    ~~~~~~~~~~~~~~~~~~~~~~~~~~~~~~~~~~~~~~~~~~~~$O_{upper}=(O_{x_{upper}},O_{y_{upper}}),O_{lower}=(O_{x_{lower},O_{y_{lower}}})$\\
    ~~~~~~~~~~~~~~~~~~~~~~~a. ~~~If $O_{upper}$and$O_{lower}$ both exist:\\
    ~~~~~~~~~~~~~~~~~~~~~~~~~~~~~~Calculating the upper/lower boundary functions, getting the Gradient Band function $F_{GB}(x,y)$\\
    ~~~~~~~~~~~~~~~~~~~~~~~~~~~~~~~~~~~~~~~~~~~~~~~~~~$\left\{
  \begin{aligned}
  f(x,y)_{upper}=y-(U + &a_0+a_1x+...+a_kx^k)\\
  f(x,y)_{lower}=y-(L + &a_0+a_1x+...+a_kx^k)\\
  X_L < x < X_{max} &, Y_L < y <Y_{max}
  \end{aligned}
  \right.$\\
    ~~~~~~~~~~~~~~~~~~~~~~~b. ~~~If only one of $O_{upper}$and$O_{lower}$ exist:\\
    ~~~~~~~~~~~~~~~~~~~~~~~~~~~~~~Calculating the upper/lower boundary functions with $(X_{max},0)$ or $(0,Y_{max})$ getting the Gradient\\
    ~~~~~~~~~~~~~~~~~~~~~~~~~~~~~~Band function $F_{GB}(x,y)$\\
    ~~~~~~~~~~~~~~~~~~~~~~~~~~~~~~~~~~~~~~~~~~~~~~~~~~$\left\{
  \begin{aligned}
  f(x,y)_{upper}=\min\{f(&X_{max},0),f(0,Y_max)\}\\
  f(x,y)_{lower}=y-(L + &a_0+a_1x+...+a_kx^k)\\
  X_L < x < X_{max} &, Y_L < y <Y_{max}
  \end{aligned}
  \right.$\\
    3. ~~~Obstacle function set generation\\
    ~~~~~~~~i. ~~~Setting $Y=[1,2,\cdots,Y_{max}]$ and $X=[1,2,\cdots,X_{max}]$ to $F_{GB}(x,y)$ respectively, generating the obstacle function\\
    ~~~~~~~~~~~~~~set $O_{baffle}=\{F_{Y_1},\cdots,F_{X_1},\cdots\}$\\
     \hline
   \end{tabular}
\end{table}
\subsection{`` 1:N '' Gradient Band-based Adversarial Training}
As the defense against adversarial attacks mainly including three directions (\cite{AkhtarN2018Threat}), which are modify the training or input data, modify the neural network, and utilize the additional model. For the last two directions, we need to modify the structure or parameters of neural network. While for traditional adversarial training under the first direction, it requires the training is performed using strong attacks and the architecture of network needs to be sufficiently expressive. Moreover, the traditional adversarial training necessitates increased training data size, which will greatly increase the total training time.

\begin{table}
\scriptsize
\centering
\begin{tabular}{l}
  \hline
  \textbf{Algorithm 2} Dominant Adversarial Attack Effectiveness Validation \\
  \hline
  Input : $Example=\{Example_1,Example_2,\cdots,Example_N\}$, $Example_i=\{T_{total_i},A_{actual_i}=(X_i,Y_i)\}$,\\ ~~~~~~~~~$D_{destination}$ \\
  Output : $Example_{attack}=\{Example_1,Example_2,\cdots,Example_m\}$\\
  ~~i. ~~~for $Example_i$ in $Example$\\
  ~~~~~~~~~~$F_{attack_i}(A_{actual_i},D_{destination},T_{total_i})$\\
  ~~~~~~~~~~~~~~~~~~~~~~~~~~~~~~~~~~~~~~~~~~~~~~~~~~~~~~=$\omega_1Relu(T_{total}-\varepsilon)+\omega_2\sqrt{\sum^n_{i=1}(A_{actual_i}-D_{destination_i})^2}$\\
  ~~~~~~~~~~if $F_{attack_i}>0$\\
  ~~~~~~~~~~~~~~then $Example_{attack}$.append($Example_i$)\\
  ~ii. ~~~ print $Example_{attack}$\\
  \hline
\end{tabular}
\end{table}
\begin{figure}
  \centering
  \includegraphics[width=16cm]{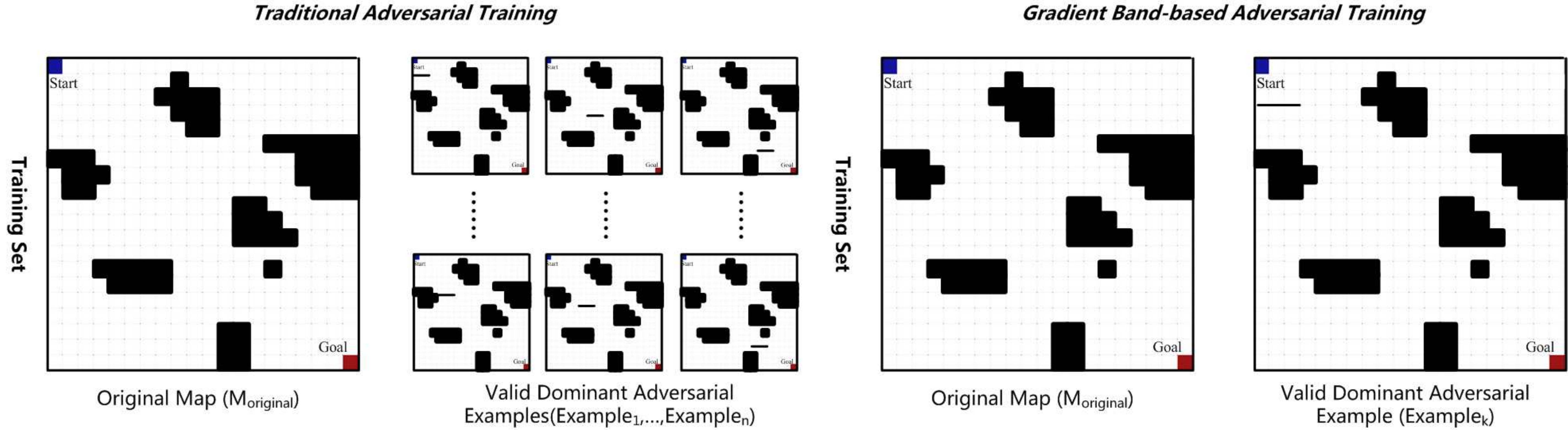}\\
  \caption{Example of the training data set for Traditional Adversarial Training and Gradient Band-based Adversarial Training. Compared with the traditional method, Gradient Band-based Adversarial Training proposed in this paper, utilizes a single example which selected randomly to realize the general immunization, which reduces the size of training data and improves the training efficiency.}
  \label{Fig:gradientband}
\end{figure}

Therefore, under such scenario, we design a defense method which can be called as \emph{Gradient Band-based Adversarial Training}. This method does not need to modify the parameters for neural network, or add additional networks. For Gradient Band-based Adversarial Training proposed in this paper, we select a single example in valid sample space $V_{Map_{N\times N}}$ randomly, to realize the ``1:N'' attack immunization for generalized dominant adversarial attack. In other words, this method utilizes a single example to conduct adversarial training, instead of using a large amount of training data under traditional method (e.g. Figure~\ref{Fig:gradientband}), which is a more concise and efficient method compared with other defense approaches.

\section{Experimental Evaluation}
In this section, we discuss the experiment results and give a comprehensive evaluation of Gradient Band-based Generalized Attack Immune model proposed in this paper.
\subsection{Experimental Setup}
\subsubsection{Experimental Preparation}
\begin{itemize}
  \item Experimental Environment
\end{itemize}

Our experiment is conducted on the operating system of Linux on the hardware environment \emph{Intel(R) Core(TM) i5-6400 CPU @2.70GHz, 8GB RAM and IT hard disk}, which can be concluded in Table. We utilize the programming language \emph{python 2.7} to implement A3C path finding under the given map.

\begin{center}
\begin{tabular}{c|l}
  \hline
  \rowcolor{mygray}
  ~~\emph{Operation System}~~ & Linux 16.04 \\
  ~~\emph{Processor}~~ & Intel(R) Core(TM) i5-6400 CPU @2.70GHz~~~~~~ \\
  \rowcolor{mygray}
  \emph{RAM} & 8GB RAM \\
  \emph{Driver} & IT hard disk \\
  \rowcolor{mygray}
  \emph{Language} & Python 2.7 \\
  \hline
\end{tabular}
\end{center}

\begin{itemize}
  \item Experimental Data
\end{itemize}

The map for automatic path finding is a grid-world of size $N\times N$ with randomly placed obstacles, in which the start position is denoted by a blue square, the destination position is denoted as a red square, and the agent for training is represented by a blue circle ( e.g. Figure \ref{Fig:exampleforDAE} ). We select 10 different map sizes ($Map=\{Map_{10\times 10},Map_{20\times 20},\cdots,Map_{100\times 100}\}$) which each with 1000 randomly generated map samples, and utilize a total of 10,000 map samples as the data base for our experiment.

\begin{itemize}
  \item Baseline Method
\end{itemize}

As the research objective in this paper, is the physical-level perturbations under A3C path finding. Hence, under such scenario, to verify the effectiveness of the generalized attack immune model proposed in this paper, we select the \emph{traditional adversarial training} and the \emph{gradient regularization} ( discussed in section 2) as the baseline method for our analysis.

\subsubsection{Experimental Analysis}

\textbf{\emph{A. Generate precision analysis for CDG algorithm}}

In order to analyze the generate precision for CDG algorithm in \emph{Generation Module} proposed in this paper, we utilize CDG algorithm to generate dominant adversarial examples for each original clean map. For the examples obtained, we utilize the well trained A3C agent to verify whether the example can attack path finding successfully. For each Example, we calculate the success criteria $F_{attack}$, to take statistics for the number of examples whose $F_{attack}>0$, then we can calculate the generation precision ($Precision_{generation}^N$) for CDG algorithm under this map.

\textbf{\emph{B. Immune precision analysis for Gradient Band-based Adversarial Training}}

We utilize a single dominant adversarial example in the valid sample space $V_{Map_{N\times N}}$, to retrain the A3C agent by Gradient Band-based Adversarial Training, then we can obtain a new A3C $agent_{new}$ with general immune ability. In order to verify this ability, we utilize all remaining examples in $V_{Map_{N\times N}}$ to conduct attack experiments on $agent_{new}$. By taking statistics for the number of dominant adversarial examples, which can be successfully immunized by $agent_{new}$ ($F_{attack} = 0$), we can measure the immune precision of the Gradient Band-based Adversarial Training proposed in this paper.

\textbf{\emph{C. Impact factors analysis for generation/immune precision}}

From the data level, we analyze the influence of map size on the generation/immune precision. We select 10 different map sizes ($Map=\{Map_{10\times 10},Map_{20\times 20},\cdots,Map_{100\times 100}\}$), under each category we generate 1,000 map samples randomly, and utilize a total of 10,000 map samples as the data base for our experiment. Calculating the generation/immune precision under each map dimension category respectively, and taking analysis to the result.

\textbf{\emph{D. Time efficiency analysis}}

Time complexity reflects the magnitude of the increase for the algorithm execution time as the input scale grows, hence, it can measures the performance of the algorithm to a large extent. Therefore, we analyze the time complexity of CDG algorithm proposed in this paper, to evaluate the efficiency based on the algorithm level. In addition, we also record the total training time for the Traditional Adversarial Training and the Gradient Band-based Adversarial Training respectively. Analyzing the time efficiency of Gradient Band-based Adversarial training proposed in this paper compared with the traditional method, to take evaluation on application level.

\textbf{\emph{E. Failed attack/immune examples analysis}}

For the CDG algorithm proposed in this paper, we take analysis of the failed attack examples which can not attack successfully, to explain the reason of failure for such dominant adversarial attack. Moreover, we try to find out the underlying rules for the failed dominant adversarial examples. In addition, for the Gradient Band-based Adversarial Training proposed in this paper, we also take analysis of the failed immune examples, which can attack the retrained A3C $agent_{new}$ after adversarial training successfully. To find the underlying rules for such examples.
\begin{table}
\scriptsize
\centering
\caption{Statistics on the attack effect for samples of Dominant Adversarial Examples}\label{Tab:attackeffect}
\begin{tabular}{|c|c|c|c|c|}
  \hline
  Original map & Dominant Adversarial Example & Time & Actual Arrived Point & $F_{attack}$ \\
  \hline
  \multirow{4}*{$Map_{A}$} & \cellcolor{mygray}$Example_{\uppercase\expandafter{\romannumeral1}}$ & \cellcolor{mygray}$T_{max} (s)$& \cellcolor{mygray}(1,2) & \cellcolor{mygray}25.07 \\
  ~ & $Example_{\uppercase\expandafter{\romannumeral2}}$ & $T_{max} (s)$ & (14,15) & 9.80 \\
  ~ & \cellcolor{mygray}$Example_{\uppercase\expandafter{\romannumeral3}}$ & \cellcolor{mygray}$T_{max} (s)$ & \cellcolor{mygray}(8,9) & \cellcolor{mygray}16.18 \\
  ~ & $Example_{\uppercase\expandafter{\romannumeral4}}$ & $T_{max} (s)$ & (6,6) & 18.79 \\
  \hline
  \multirow{4}*{$Map_{B}$} & \cellcolor{mygray}$Example_{\uppercase\expandafter{\romannumeral1}}$ & \cellcolor{mygray}$T_{max} (s)$ & \cellcolor{mygray}(15,3) & \cellcolor{mygray}17.37 \\
  ~ & $Example_{\uppercase\expandafter{\romannumeral2}}$ & $T_{max} (s)$ & (17,9) & 12.65 \\
  ~ & \cellcolor{mygray}$Example_{\uppercase\expandafter{\romannumeral3}}$ & \cellcolor{mygray}94 (s) & \cellcolor{mygray}(19,19) & \cellcolor{mygray}16.00 \\
  ~ & $Example_{\uppercase\expandafter{\romannumeral4}}$ & $T_{max} (s)$ & (15,2) & 18.10 \\
  \hline
  \multirow{4}*{$Map_{C}$} & \cellcolor{mygray}$Example_{\uppercase\expandafter{\romannumeral1}}$ & \cellcolor{mygray}$T_{max} (s)$ & \cellcolor{mygray}(13,8) & \cellcolor{mygray}14.40 \\
  ~ & $Example_{\uppercase\expandafter{\romannumeral2}}$ & 85 (s) & (19,19) & 13.75 \\
  ~ & $\cellcolor{mygray}Example_{\uppercase\expandafter{\romannumeral3}}$ & \cellcolor{mygray}$T_{max} (s)$ & \cellcolor{mygray}(16,16) & \cellcolor{mygray}8.18 \\
  ~ & $Example_{\uppercase\expandafter{\romannumeral4}}$ & $T_{max}(s) $ & (6,1) & 21.65 \\
  \hline
  \multirow{4}*{$Map_{D}$} & \cellcolor{mygray}$Example_{\uppercase\expandafter{\romannumeral1}}$ & \cellcolor{mygray}$109 (s)$ & \cellcolor{mygray}(19,19) & \cellcolor{mygray}19.75 \\
  ~ & $Example_{\uppercase\expandafter{\romannumeral2}}$ & $T_{max}$ & (6,0) & 17.27 \\
  ~ & \cellcolor{mygray}$Example_{\uppercase\expandafter{\romannumeral3}}$ & \cellcolor{mygray}$8 (s)$ & \cellcolor{mygray}(19,19) & \cellcolor{mygray}0.00 \\
  ~ & $Example_{\uppercase\expandafter{\romannumeral4}}$ & $T_{max}(s) $ & (19,11) & 6.00 \\
  \hline
\end{tabular}
\end{table}
\subsection{Experimental Result and Evaluation}
\subsubsection{Dominant Adversarial Examples Generation Validity Analysis}
In order to verify the generation effectiveness of CDG algorithm proposed in this paper, we take analysis of the generation precision.

\begin{figure}
  \centering
  \includegraphics[width=14cm]{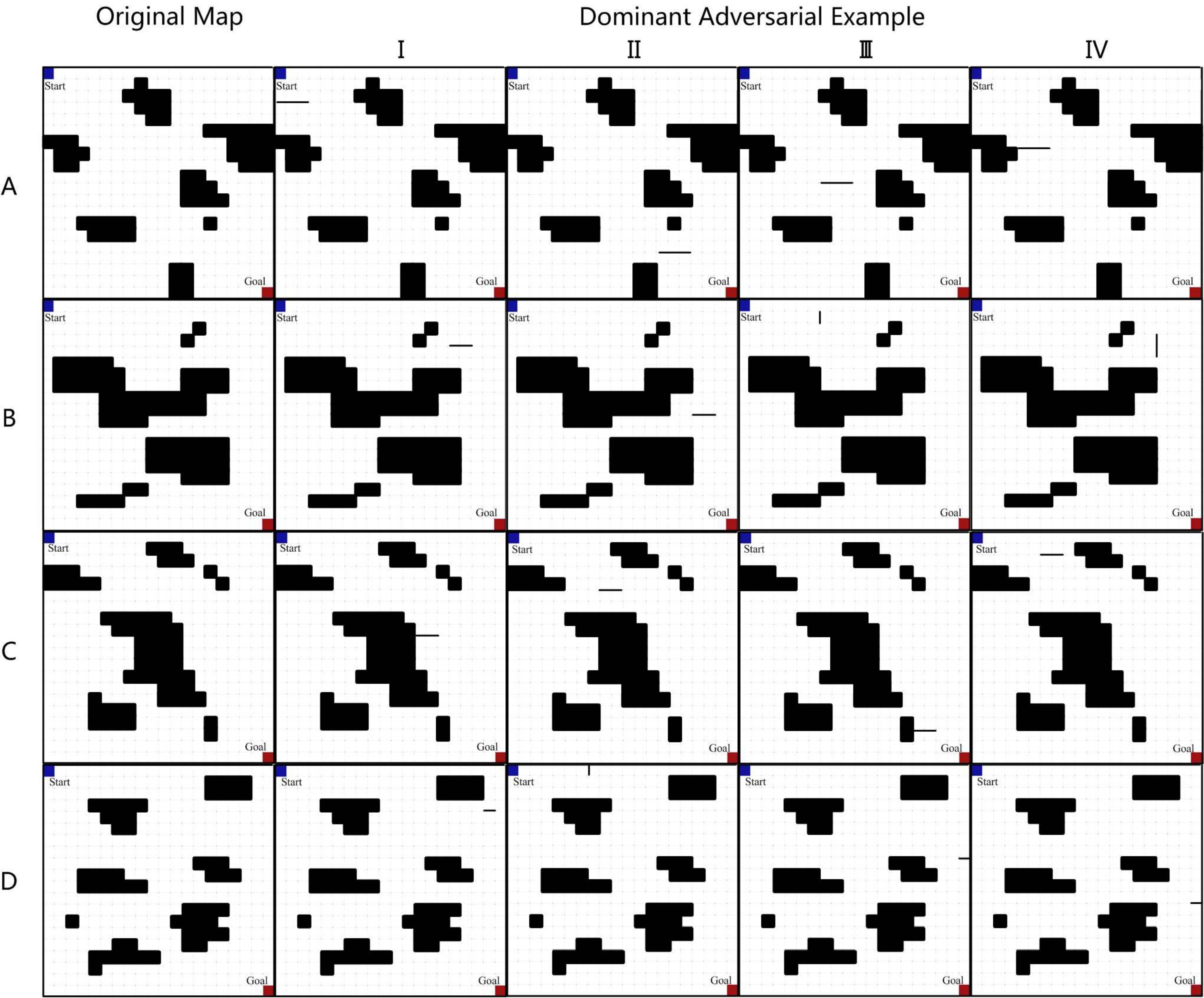}\\
  \caption{Samples for Dominant Adversarial Examples. For the first column is the original clean map for path finding. Four columns on the right are the samples for Dominant Adversarial Examples generated by CDG algorithm proposed in this paper under \emph{Generation Module}.}
  \label{Fig:DAEvalidation}
\end{figure}

For a original clean $Map_{N\times N}$, we utilize CDG algorithm proposed in this paper under Generation Module, to generate dominant adversarial examples. As shown in Figure\ref{Fig:DAEvalidation}, is the samples for dominant adversarial examples generation results against original clean map shown on the left column. We calculate the $F_{attack}$ value for each example according to the attack success criterion, to verify whether this example can attack the well trained A3C agent successfully (e.g. Table\ref{Tab:attackeffect}). If $F_{attack}>0$, then denotes that this dominant adversarial example can attack A3C path finding successfully. If $F_{attack}=0$ represents that this example is an invalid adversarial example. By taking statistics of the examples whose $F_{attack}>0$ to compute the generation precision for CDG algorithm under this map, which can be concluded as

\begin{equation}
  Precision_{generation}^N = \frac{N_{success}}{M_{total}}
\end{equation}
where $N_{success}$ denotes the number of examples that can attack successfully, $M_{total}$ is the total number of examples generated by CDG algorithm, and N denotes the map size.

In order to improve the accuracy for our analysis, we calculate the generation precision for 1,000 maps under each map size, and utilize the average value of $Precision_{generation}^N$ as the generation precision under $Map_{N \times N}$. As shown in Table~\ref{Tab:g/iprecision}, for all the categories of map-size, the lowest value for $Precision_{generation}^N$ is $91.91\%$ and the highest value is $96.59\%$. Proving that the CDG algorithm proposed in our model, is able to generate dominant adversarial examples which can attack A3C path finding successfully with high confidence.

\subsubsection{Generalized Immune Effectiveness Analysis}

\begin{itemize}
  \item ``1:N'' immune precision analysis
\end{itemize}

We utilize any single example in the valid sample space $V_{Map_{N\times N}}$ which can attack A3C path finding successfully under $Map_{N\times N}$ to retain the A3C agent with Gradient Band-based Adversarial Training, then we can obtain a new $agent_{new}$ with general immune ability of ``1:N''. In order to verify the this generalized immunization ability of $agent_{new}$, we take use of all remaining examples in $V_{Map_{N\times N}}$ to attack $agent_{new}$ under A3C automatic path finding.

By calculating the $F_{attack}$ value for each example, we can take statistics of the number for dominant adversarial examples which can be immunized by $agent_{new}$ ($F_{attack}=0$), then we can calculate the immune precision to evaluate the effectiveness for Gradient Band-based Adversarial Training under $Map_{N\times N}$, which can be concluded as:
\begin{equation}
  Precision_{immnue}^N = \frac{I_{success}}{N_{success}}
\end{equation}
where $I_{success}$ denotes the number of examples which can be successfully immunized by $agent_{new}$, and $N_{success}$ is the number of examples that can attack successfully in the valid sample space $V_{Map_{N\times N}}$, and N denotes the map size.

\begin{table}
\scriptsize
\centering
\caption{Statistics on the generation/immune precision under each map size.}\label{Tab:g/iprecision}
\begin{tabular}{|c|c|c|}
  \hline
  ~~~~~~~~~~~Map Size~~~~~~~~~~~ & ~~~~~~~Generation Precision~~~~~~~ & ~~~~~~~Immune Precision~~~~~~~ \\
  \hline
  \rowcolor{mygray}
  \emph{Map $10\times 10$} & 94.44\% & 98.64\% \\
  \emph{Map $20\times 20$} & 96.59\% & 97.22\% \\
  \rowcolor{mygray}
  \emph{Map $30\times 30$} & 94.83\% & 96.36\% \\
  \emph{Map $40\times 40$} & 95.15\% & 95.67\% \\
  \rowcolor{mygray}
  \emph{Map $50\times 50$} & 94.90\% & 95.70\% \\
  \emph{Map $60\times 60$} & 93.22\% & 94.55\% \\
  \rowcolor{mygray}
  \emph{Map $70\times 70$} & 93.48\% & 93.02\% \\
  \emph{Map $80\times 80$} & 92.77\% & 94.17\% \\
  \rowcolor{mygray}
  \emph{Map $90\times 90$} & 92.70\% & 94.05\% \\
  \emph{Map $100\times 100$} & 91.91\% & 93.89\% \\
  \hline
\end{tabular}
\end{table}
\begin{itemize}
  \item Baseline method comparison
\end{itemize}
\begin{figure}
  \centering
  \includegraphics[width=16cm]{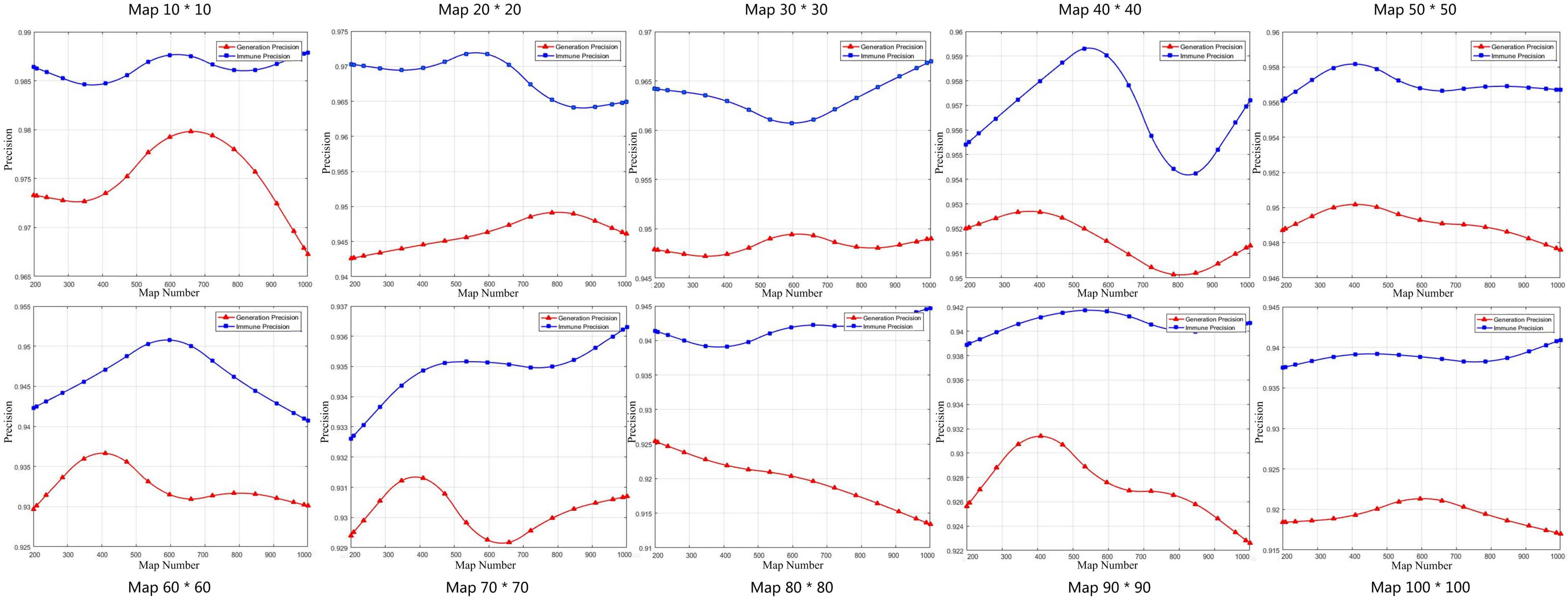}\\
  \caption{Statistics of the generation/immune precision under each map size. The 10 categories for map size can be concluded as $Map=\{Map_{10\times 10},Map_{20\times 20},\cdots,Map_{100\times 100}\}$, which can be shown in each graph. The red line denotes the precision curve for generation precision, and the blue line denotes the precision curve for immune precision.}
  \label{Fig:mapanalysis}
\end{figure}
In order to improve the accuracy of analysis, we conduct experiments on 1,000 maps under each map size, and take the average value of $Precision_{immune}^N$ as the immune precision under $Map_{N\times N}$. As shown in Table~\ref{Tab:g/iprecision}, is the immune precision under each map size, the lowest value of immune precision is $93.89\%$, and the highest value of immune precision is $98.64\%$, which can prove that Gradient Band-based Adversarial Training proposed in this paper, can makes the $agent_{new}$ trained with a single example realizing the generalized immunity to the vast majority of valid attack examples. In other words, the $agent_{new}$ generated by Gradient Band-based Adversarial Training, can realize the general adversarial examples immunization of ``1:N'' with a high confidence.

As shown in Figure~\ref{Fig:comparison} is the immune precision comparison curve, for the Gradient Band-based Adversarial Training proposed in this paper and the baseline methods. The green line denotes the baseline method for \emph{Gradient Regularization(GR)} which own the lowest precision (83.67\%), and the blue line represents the baseline method for \emph{Traditional Adversarial Training} with own the highest precision (96.54\%), and the red line is the immune precision curve for Gradient Band-based Adversarial Training proposed in this paper, which own a relatively high immune precision (95.50\%).

Although the immune precision of Gradient Band-based Adversarial Training is lower than that of Traditional Adversarial Training, however, the time processing efficiency of our method is much higher than that of the traditional one, which can be shown in the Table~\ref{Tab:efficiency}. Therefore, under such scenario, we believe that the performance of Gradient Band-based Adversarial Training proposed in this paper is the best.
\setcounter{figure}{8}
\begin{figure}[htbp!]
  \subfigure{
  \begin{minipage}[t]{0.5\linewidth}
  \centering
    \includegraphics[width = 2.5in]{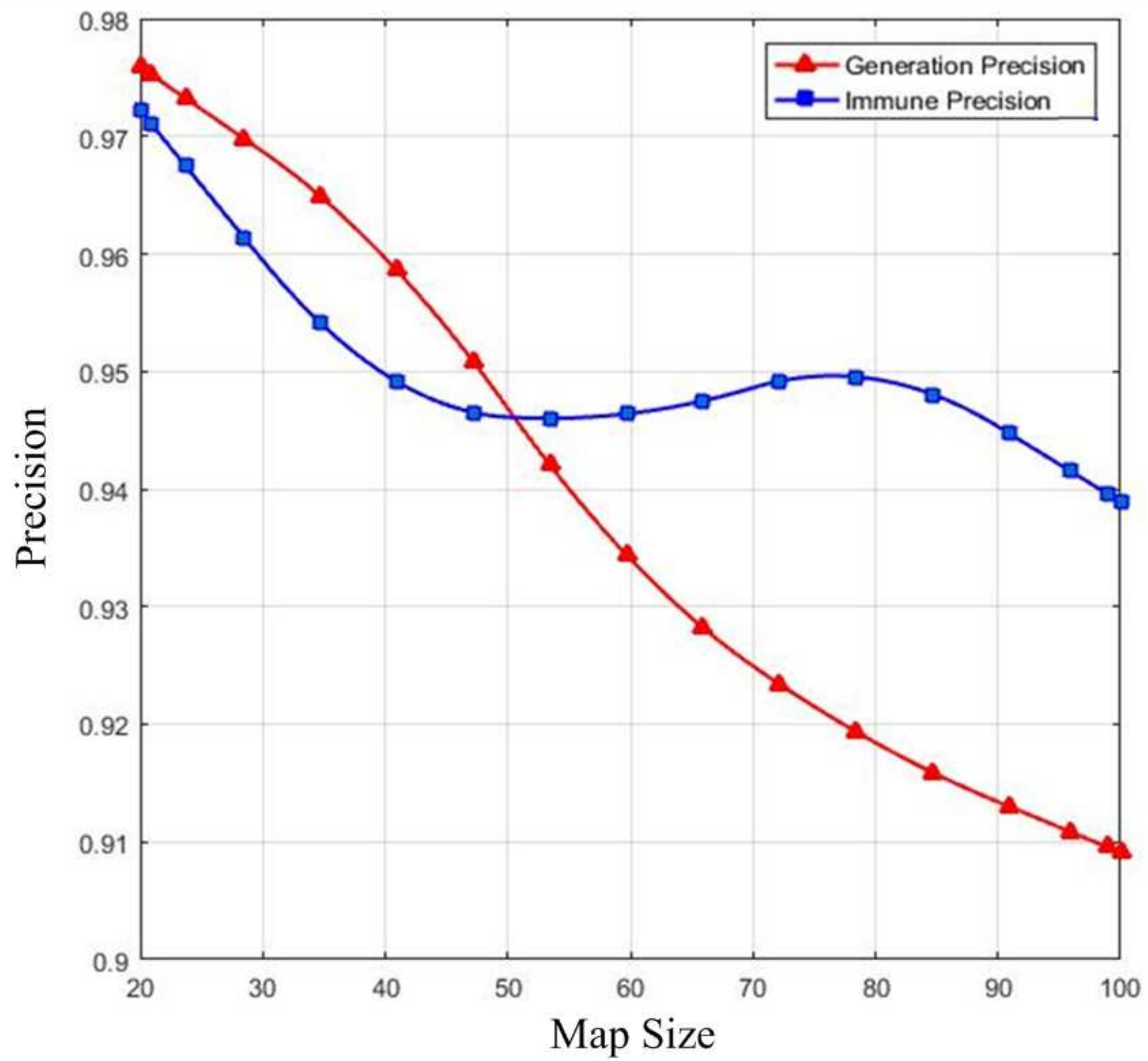}
    \caption{The effect of map-size on generation/immune precision\label{Fig:attack-defense}}
    \end{minipage}%
    \begin{minipage}[t]{0.5\linewidth}
    \centering
    \includegraphics[width = 2.5in]{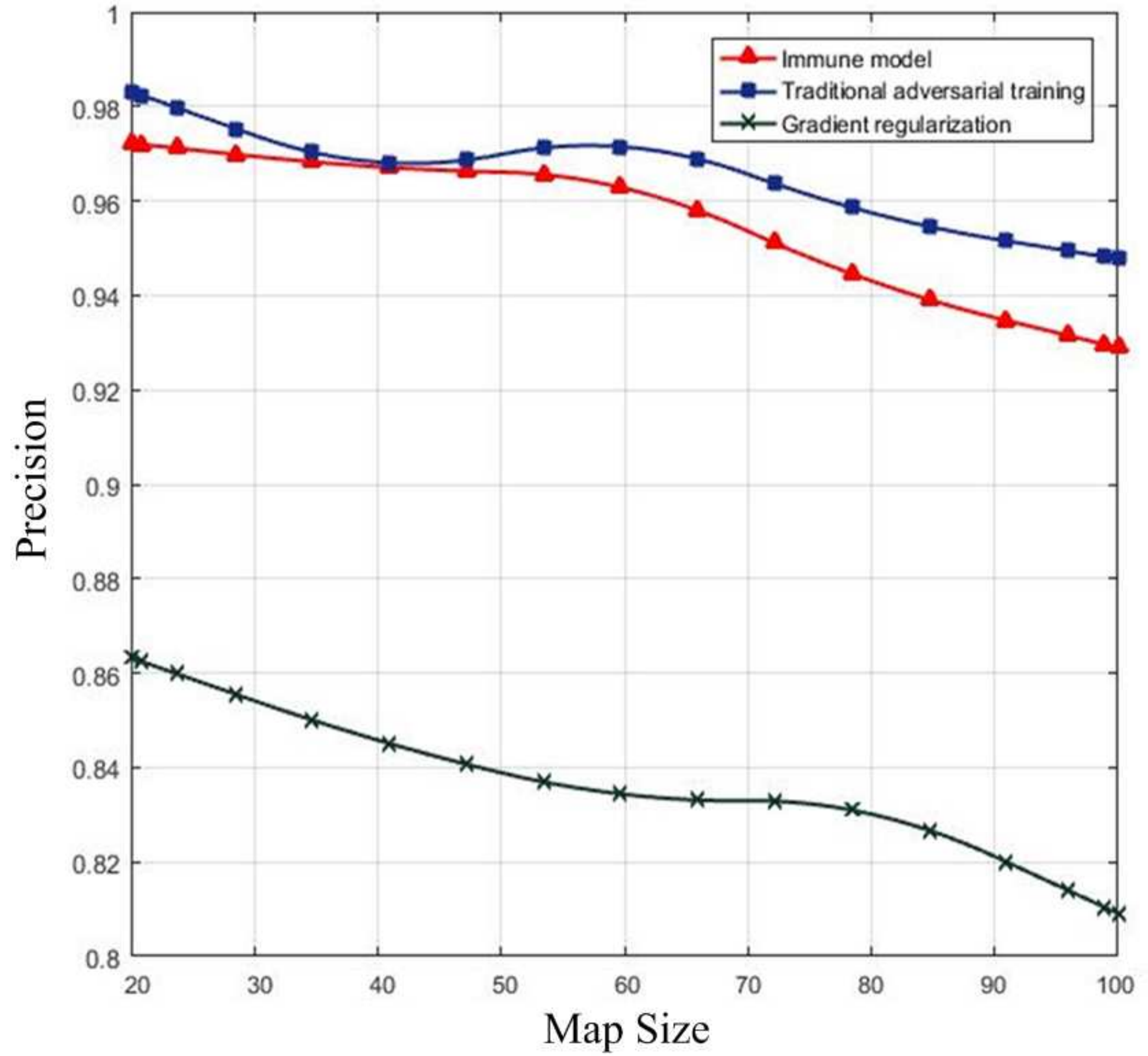}
    \caption{Immune precision comparison curve for three methods\label{Fig:comparison}}
    \end{minipage}
    }
\end{figure}

\subsubsection{Impact Factors Analysis}

Through experiments we find that, the size of map has effect on generation/immune precision. We calculate the generation/immune precision values under different categories of map size, which can be shown in Figure~\ref{Fig:mapanalysis} and~\ref{Fig:attack-defense}. The blue line denotes the distribution curve of immune precision, and the red one represents the distribution curve of generation precision. As shown in Figure~\ref{Fig:mapanalysis} is the generation/immune precision distribution of 1,000 map samples under each map size, and Figure~\ref{Fig:attack-defense} shows the change of the overall precision curve as the changing for map size.

We can see that, from the overall trend, with the increase of map size, the precision of generation and immune has decreased to some extent. However, the lowest value for generation precision and immune precision are 91.91\% and 93.89\% respectively. Therefore, it can be considered that, the Gradient Band-based Generalized Attack Immune model proposed in this paper, can realize generalized immunization to dominant adversarial attacks with a relatively high accuracy under different categories of map size.

\setcounter{figure}{11}
\begin{figure}
  \centering
  \includegraphics[width=15cm]{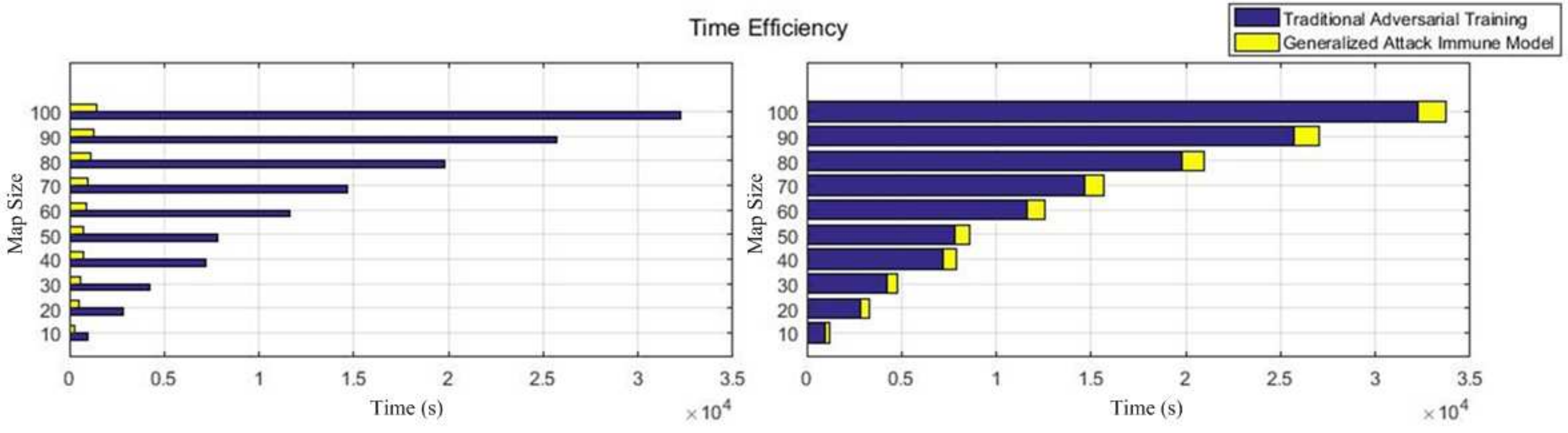}\\
  \caption{Time efficiency comparison}\label{Fig:Time}
\end{figure}
\begin{table}
\scriptsize
\centering
\caption{Time efficiency for Traditional Adversarial Training and our Immune Model}\label{Tab:efficiency}
\begin{tabular}{|c|c|c|c|c|c|c|c|c|c|c|}
  \hline
  \backslashbox{Method}{Map-size} & $Map_{10\times 10}$ & $Map_{20\times 20}$ & $Map_{30\times 30}$ & $Map_{40\times 40}$ & $Map_{50\times 50}$  \\
  \hline
  ~~~~Traditional Adversarial Training~~~~ & 984.30 s & 2,832.21 s & 4,248.63 s & 7,162.80 s & 7,818.12 s  \\
  \rowcolor{mygray}~~~Generalized Attack Immune Model~~~ & 256.13 s & 501.40 s & 547.50 s & 705.28 s & 769.52 s  \\
  \hline
  \backslashbox{Method}{Map-size} & $Map_{60\times 60}$ & $Map_{70\times 70}$ & $Map_{80\times 80}$ & $Map_{90\times 90}$ & $Map_{100\times 100}$ \\
  \hline
  ~~~~Traditional Adversarial Training~~~~ & 11,624.28 s & 14,686.98 s & 19,803.42 s & 25,727.52 & 32,259.42 s\\
  \rowcolor{mygray}~~~Generalized Attack Immune Model~~~ & 925.53 s & 981.75 s & 1,140.75 & 1,302.00 s & 1,455.75 s\\
  \hline
\end{tabular}
\end{table}
\subsubsection{Time Complexity Analysis}
\begin{itemize}
  \item Time efficiency for CDG algorithm
\end{itemize}

In order to measure the processing efficiency for CDG algorithm proposed in this paper, we take analysis of the time complexity for this algorithm to prove the time efficiency of it.

For the gradient function calculation, the time complexity for value processing is $O(N)$, and the time complexity for calculating the gradient function is $O(N)$, as the Gradient Descent algorithm used for this step only needs to calculate the first derivative. Meanwhile, for the gradient band calculation, the time complexity for fitting function calculation is $O(kN^2)$, while the time complexity for computing the distances from obstacles edge points to the gradient function is $O(N)$, and the time complexity for the calculating to upper/lower boundary functions is $O(1)$. In addition, for the obstacle function set generation, the time complexity is $O(N)$.

{
\small
\centering
\begin{tabular}{l|l|c}
  \hline

  \multicolumn{2}{c|}{Algorithm Steps} & Time Complexity \\
  \hline
  \multirow{2}*{Gradient function calculation}& \cellcolor{mygray}Value processing & \cellcolor{mygray}$O(N)$ \\
  ~& Gradient function calculation& $O(N)$ \\
  \multirow{3}*{Gradient band calculation} & \cellcolor{mygray}Fitting function calculation&\cellcolor{mygray}$O(kN^2)$ \\
  ~&Distances computation & $O(N)$ \\
  ~&\cellcolor{mygray}Upper/Lower boundary functions calculation&\cellcolor{mygray}$O(1)$\\

  \multicolumn{2}{c|}{Obstacle function set generation}&$O(N)$\\
  \hline
\end{tabular}
}

Therefore, we can see that the time complexity for CDG algorithm proposed in this paper is polynomial, which can prove that our algorithm can generate dominant adversarial examples with a high processing efficiency.

\begin{itemize}
  \item Time efficiency for Gradient Band-based Adversarial Training
\end{itemize}

In order to prove the time efficiency for the Gradient Band-based Adversarial Training proposed in this paper, we record the training time for Traditional Adversarial Training and our training method respectively, and the results can be shown in Table~\ref{Tab:efficiency} and Figure~\ref{Fig:Time}. We find that the training method proposed in this paper is much better than the Traditional Adversarial Training in processing efficiency. Therefore, it is reasonable to use the Gradient Band-based Adversarial Training in our experiment for immune training.

\subsubsection{Failed Examples Analysis}
\begin{itemize}
  \item Failed attack examples analysis
\end{itemize}

\begin{figure}
  \centering
  \includegraphics[width=12cm]{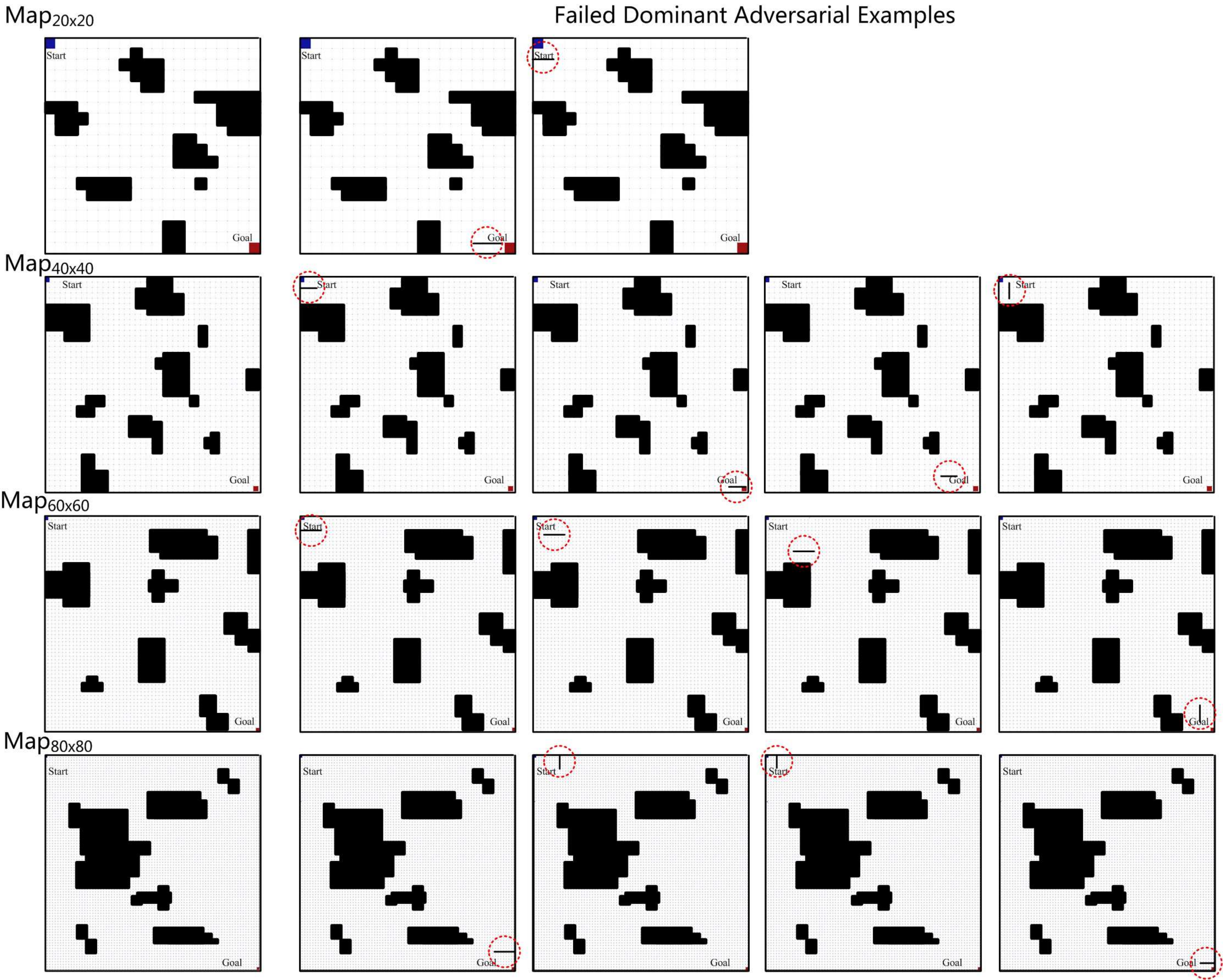}\\
  \caption{The samples for failed attack examples. By taking analysis, we can find that adding perturbations to the area where close to the start point and destination point, will reduce the effectiveness of attack.}
  \label{Fig:failattack}
\end{figure}

We take analysis of the failed attack examples generated by CDG algorithm, which can be shown in Figure~\ref{Fig:failattack}. We choose samples under four categories of map size, which are $Map_{20\times 20}$,$Map_{40\times 40}$,$Map_{60\times 60}$,$Map_{80\times 80}$, and it is worth to be noted that under $Map_{20\times 20}$ we only discover two invalid dominant adversarial examples for attacking, while under other map dimensions, we choose four invalid examples randomly to take analysis.

By taking analysis to the failed dominant adversarial examples, we can find that, the interference obstacles added are close to the starting point or the destination point. When the ``physical-level'' perturbations added to such area, the dominant adversarial examples generated by CDG will can not attack A3C path finding effectively.

\begin{itemize}
  \item Failed immune examples analysis
\end{itemize}

We also take analysis of the failed immune examples, which can still attack the $agent_{new}$ that retrained by Gradient Band-based Adversarial Training. As shown in Figure is the samples for failed immune examples. We choose the same original map with the analysis of failed attack examples to make a better comparison, and for $Map_{20\times 20}$ we only discover a single failed example which can attack $agent_{new}$ successfully, meanwhile, for $Map_{40\times 40}$ three failed immune examples have been found.

By taking analysis to the failed immune examples, which can still attack $agent_{new}$ that retrained by Gradient Band-based Adversarial Training, we discover that, most of the perturbations for failed examples are in the area, with high density of original obstacles on the Gradient Band. In other words, when adding perturbations to such area will reduce the effectiveness of immunization.

\begin{figure}
  \centering
  \includegraphics[width=12cm]{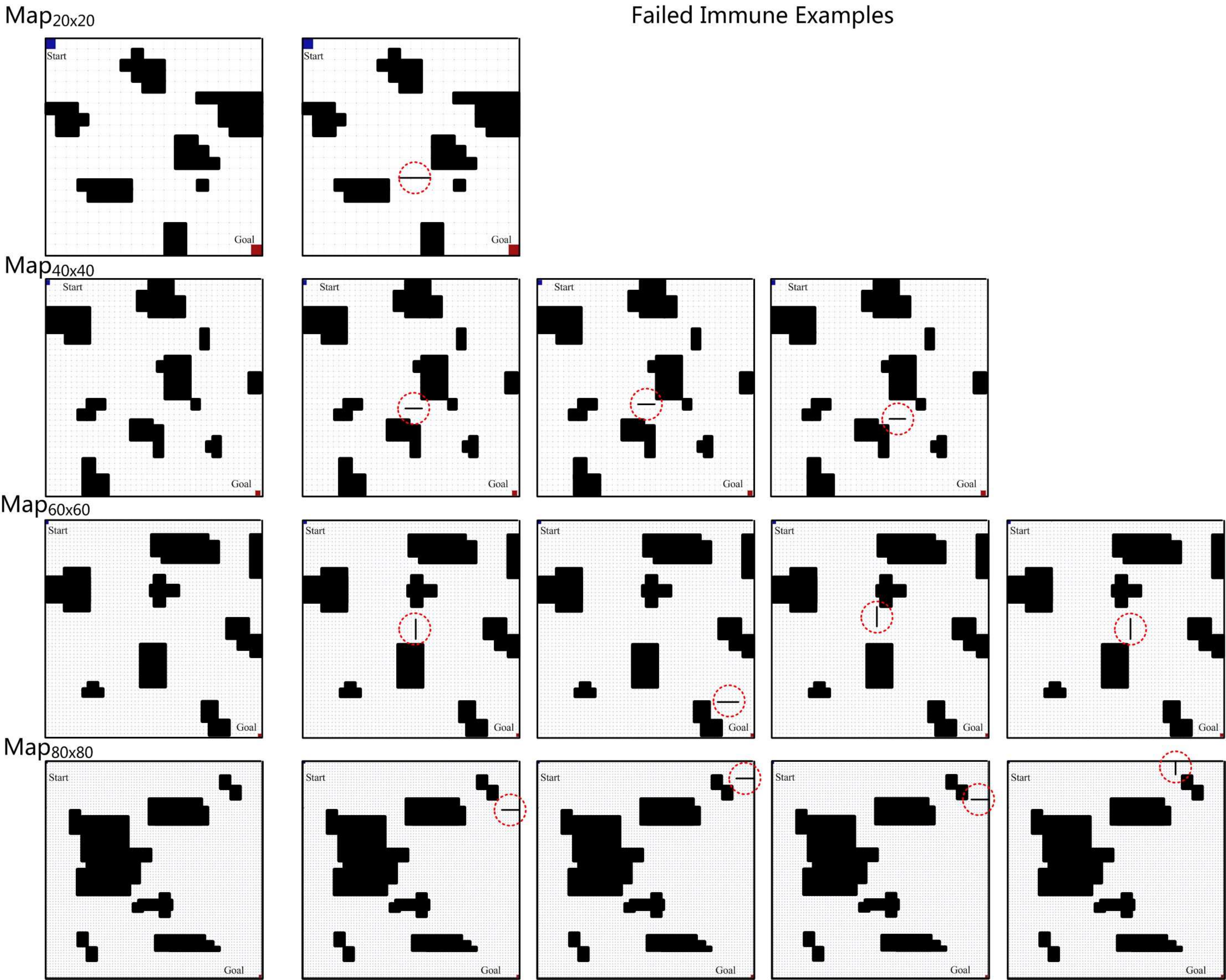}\\
  \caption{The samples for failed immune examples. By taking analysis, we can find that adding perturbations to the area with high density of original obstacles on the Gradient Band, will reduce the effectiveness of immunization.}
  \label{Tab:failimmune}
\end{figure}

\section{Conclusion}
In this paper, we pay attention to the adversarial research on automatic path finding, and show that dominant adversarial examples are effective when targeting A3C path finding. Moreover, we design a Common Dominant Adversarial Examples Generation Method (CDG), which can generate dominant adversarial examples for any given map. In addition, we propose a Gradient Band-based Adversarial Training, which trained with a single randomly choose dominant adversarial example (`` 1 '') without taking any modification, to realize the ``1:N'' attack immunity for generalized dominant adversarial examples (`` N '').

We provide extensive experimental evidence on 10,000 map samples generated randomly under 10 different categories of map size. The lowest generation precision for CDG algorithm is $91.91\%$, and the highest precision is $96.59\%$, which show that our CDG algorithm can generate valid dominant adversarial examples with a high confidence. Moreover, as the lowest immune precision is $93.89\%$, and the highest precision is $98.64\%$, the Gradient Band-based Adversarial Training proposed in this paper, can realize the ``1:N'' attack immunity for generalized dominant adversarial examples with high immune precision.





\section*{Acknowledgements}
This work was supported in part by the Natural Science Foundation of China under Grants 61672092, in part by the Fundamental Research Funds for the Central Universities of China under Grants 2018JBZ103.

\end{document}